\documentclass{article}

\usepackage{arxiv}

\usepackage[T1]{fontenc}    
\usepackage{hyperref}       
\usepackage{url}            
\usepackage{booktabs}       
\usepackage{amsfonts}       
\usepackage{nicefrac}       
\usepackage{microtype}      
\usepackage{lipsum}
\usepackage{graphicx}

\usepackage{times}
\usepackage{soul}
\usepackage{url}
\usepackage[small]{caption}

\usepackage{float}
\usepackage{cite}
\usepackage{subfigure}
\usepackage{caption}
\usepackage{multirow}
\usepackage{enumitem}

\begin{document}
\title{A Deep Learning Framework for Generation and Analysis of Driving Scenario Trajectories\thanks{Andreas Demetriou, Henrik Alfsvåg, Sadegh Rahrovani and Morteza Haghir Chehreghani have contributed equally to this work. The work is published in \emph{SN Computer Science, 4, 251, 2023.}}}

\author{Andreas Demetriou \\
        Chalmers University of Technology\\
        SE-412 96 Gothenburg, Sweden\\
        \And
        Henrik Alfsvåg \\
        Chalmers University of Technology\\
        SE-412 96 Gothenburg, Sweden\\
        \And
        Sadegh Rahrovani \\
        Autonomous Drive Department, Volvo Cars \\
        Gothenburg, Sweden
        \And
        Morteza Haghir Chehreghani  \\
        Chalmers University of Technology\\
        SE-412 96 Gothenburg, Sweden\\
        Email: \texttt{morteza.chehreghani@chalmers.se}
}

\maketitle

\begin{abstract}
We propose a unified deep learning framework for generation and analysis of driving scenario trajectories, and validate its effectiveness in a principled way. In order to model and generate scenarios of trajectories with different lengths, we develop two approaches. First, we adapt the Recurrent Conditional Generative Adversarial Networks (RC-GAN) by conditioning on the length of the trajectories. This provides us the flexibility to generate variable-length driving trajectories, a desirable feature for scenario test case generation in the verification of autonomous driving. Second, we develop an architecture based on Recurrent Autoencoder with GANs in order to obviate the variable length issue, wherein we train a GAN to learn/generate the latent representations of original trajectories. In this approach, we train an integrated feed-forward neural network to estimate the length of the trajectories to be able to bring them back from the latent space representation.
In addition to trajectory generation, we employ the trained autoencoder as a feature extractor, for the purpose of clustering and anomaly detection, in order to obtain further insights into the collected scenario dataset. We experimentally investigate the performance of the proposed framework on real-world scenario trajectories obtained from in-field data collection.

\end{abstract}

\section{Introduction}
\begin{figure}[t]
    \centering
    \includegraphics[width=0.65\textwidth]{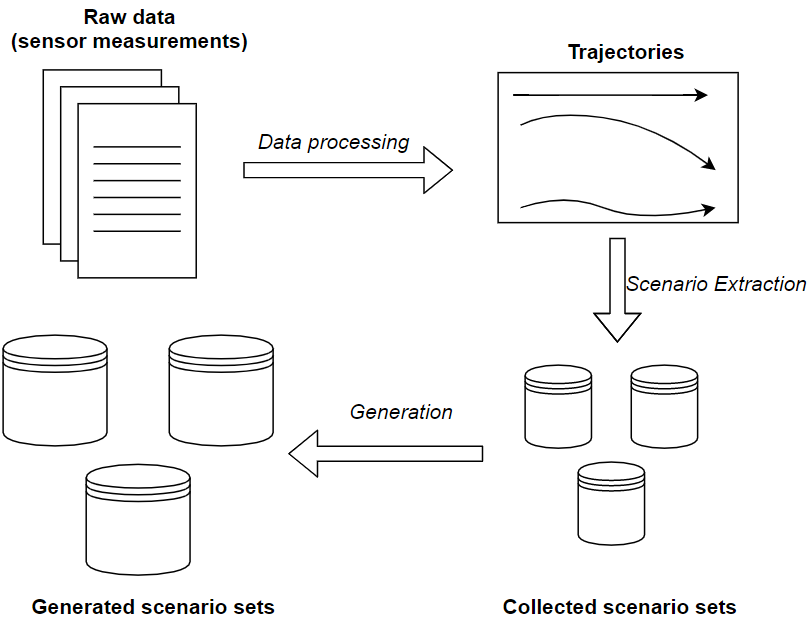}
    \caption{The full workflow from raw data to scenario database.}
    \label{full-workflow}
\end{figure}

The future of transportation is tightly connected to Autonomous Driving (AD).
While a lot of progress has been made in recent years in these areas, there are still obstacles to overcome. One of the most critical issues is the safety verification of AD. In order to assess with confidence the safety of AD, statistical analyses have shown that fully autonomous vehicles would have to be driven for hundreds of millions of kilometers \cite{kalra2016driving}. This is not feasible, particularly in cases when we need to assess different system design proposals or in case of system changes, since the same amount of distance needs to be driven again by the AD vehicle for the verification sign-off. Thus, a data-driven scenario-based verification approach that shifts performing tests in the fields to a virtual environment provides a systematic approach to tackle safety verification. This approach requires a scenario database to be created by extracting driving scenarios (e.g. cut-in, overtaking, etc.) that the AD vehicle is exposed to in naturalistic driving situations. Scenarios are obtained through time series (sequence of the ego-vehicle states and the surrounding objects) which in turn are the processed data collected by sensors of the AD vehicle. Once such a scenario database is developed, it can be used for test case generation and verification of the AD functionality in a virtual environment \cite{kim2016testing}. Note that, scenario extraction can, in general, be addressed with two approaches: an explicit rule-based approach \cite{zhao2017trafficnet} (that requires expert domain knowledge) and a (machine learning based) clustering approach \cite{wang2020clustering, Wang_2018, martinsson2018clustering, li2006coarse}, where they can complement each other. Fig.\ref{full-workflow} illustrates the high-level overview of the full process from the raw logged data to the scenario database with a sufficient number of scenarios for verification.

However, several challenges should be addressed in order to create a reliable scenario database. First, a huge amount of data is still needed to be collected and processed in order to build such a scenario database. In particular, the existing data might be imbalanced or insufficient. Second, in order to assure safety in vehicles, AD functionality needs to pass safety tests not only based on “real” scenarios (also called test cases) collected from field driving tests, but also based on many perturbed (similar) trajectories that might have not been collected in real driving data collection. To address these issues, building generative models (by mimicking the variation available in the collected scenario data) to create realistic synthetic scenarios is a main focus of this work.

Thereby, we propose a unified deep learning framework for generation and analysis of driving scenario trajectories, and validate its effectiveness in a principled way. We investigate the performance of different variants of Generative Adversarial Networks (GANs) \cite{goodfellow2014generative} for generating driving trajectories. GANs have shown promising results in several tasks related to the generation of synthetic data. In this paper, since the data is sequential, we employ recurrent architectures to extract the sequential nature of data. The first approach consists of a recurrent GAN (without an autoencoder). We adapt the Recurrent Conditional Generative Adversarial Networks (RC-GAN) by conditioning on the length of the trajectories. This provides us the flexibility to generate variable-length driving trajectories, a desirable feature for scenario test case generation in AD verification. The second approach consists of a recurrent autoencoder and a GAN for learning/generating  latent-space representations of trajectories of different lengths. In this approach, it is essential to know the length of the trajectories in order to bring them back from the latent space representation. We overcome this issue by training an integrated feed-forward neural network to estimate the lengths based on the latent space representations.

At the same time, the recurrent autoencoder can be used as a feature extractor. Thus, we analyze such latent space features in the context of exploratory data analysis in order to obtain further insights into the collected scenario set via clustering and anomaly detection.
As mentioned earlier, clustering can be useful for scenario extraction, as an alternative solution to explicit rule-based methods that might be subject to misspecification. Clustering can also provide an effective tool for data visualization and exploration.
We demonstrate the performance of the framework on real-world in-field scenario trajectories collected by Volvo Cars Corporation (VCC).

This work is  an extension of our publication in \cite{DemetriouARC20}. The extension includes different aspects such as i) further elaboration of the methods on trajectory generation using GANs, ii) a clustering  method consistent with the proposed deep learning framework, in particular with the respective latent representation, iii) an outlier detection mechanism of the trajectories based on the latent space representation using the developed recurrent autoencoder, iv) discussion on the applicability of the proposed clustering and outlier detection mechanisms for Autonomous Driving applications, and v) novel experimental studies and investigations, in particular for the clustering and outlier detection components.

\section{Background}
\subsection{Problem Description}
We are provided with the data collected by Volvo Cars Corporation. This dataset consists of information about the ego vehicle and its surroundings such as detected objects, road conditions, etc. We focus on generating realistic scenario trajectories, in particular, the cut-in trajectories for a specific tracked vehicle, and their analysis in the context of exploratory data analysis. To describe a trajectory, we consider two features: the relative lateral and longitude positions of the vehicle with respect to the ego vehicle.

To generate and analyze trajectories, our framework performs the following steps.

\begin{itemize}[leftmargin=*]
  \item Extract scenarios from the logged data, which is done with explicit rules defined by an expert.
  Note that all surrounding/target cars in the field of view (of the ego car), and the lane marking signals are available. So the rule-based scenario functions work based on this information and they assign a start time stamp and end timestamp to a scenario (e.g.,  start the cut-in scenario a couple of seconds before the target car passes the lane marking and enters the ego car's lane, and stop cut-in after the target car becomes the lead car in front of the ego car). This will be discussed more in the next section, Scenario Extraction.
  \item Build the generative models for synthesizing/generating trajectories similar to the ones collected from the in-field test.
  \item Evaluate the obtained results and compare the generated trajectories versus the real ones. This step is done by visual inspection and the metrics that will be introduced.
\end{itemize}

Besides the explicit rule-based approach for scenario extraction, a clustering method can be used as well. Clustering has some advantages. Firstly, it enables one to detect scenarios that lie on the border of two scenario classes and thus finds more complex driving patterns/scenarios. Second, explicit rules could miss outliers. Moreover, explicit rules require expert domain knowledge and a hard threshold to define scenarios, which is nontrivial to formulate and calibrate when the dimensionality of data increases. Thus, clustering, when used in combination with an explicit rule-based approach, provides exploratory insights from the data and is suitable w.r.t. scalability. Also, the labels provided by the explicit rule-based approach can be verified by the clustering-based approach for consistency, where the false positive/negative cases can be investigated further by camera sensors video check. Calibration of scenario definition threshold could be done afterwards, when these valuable miss-classified labels have been investigated. This consistency check between the two approaches can accelerate the label verification process considerably since only a limited number of video checking might be required.

\subsection{Related Work}
\subsubsection{Generation}
One approach to generate driving trajectories is based on simulations of physical models, including the vehicle dynamics and driver model. This is a promising approach, but it needs to be used in combination with other solutions, as validating those simulation models is as challenging as the verification of the AD problem. Also, the simulation of high-fidelity models can be computationally demanding w.r.t. computational and storage resources.

GANs \cite{goodfellow2014generative} are the most popular paradigms for synthetic data generation in the context of modern deep neural networks.
They have been employed and developed in several applications such as image processing, computer vision, text generation, natural language processing and translation \cite{KurachLZMG19,HwangJY19,NIPS2019_8315,NIPS2019_8682,ZhangXLZWHM19}.

A related work has been developed based on generating errors for the sensors' measurements using recurrent conditional GANs \cite{arnelid2019recurrent}. This method can be used to make simulated data look more realistic.
The study in \cite{krajewski2018data} considers the rather similar problem of maneuver modeling  with InfoGAN and $\beta$-VAE. These generative models show satisfactory results. However, the data in this work is collected by a drone which we consider to be a limitation.
\cite{ding2018new} presents `Multi-vehicle Trajectory Generator' (MTG) that is an improved version of $\beta$-VAE with recurrent networks. Moreover, it shows that the proposed MTG produces more stable results than InfoGAN or $\beta$-VAE.

\subsubsection{Clustering}
Several methods have been proposed for clustering based on time series and trajectory analysis \cite{liao2005clustering, Chehreghani16}, in particular for vehicle trajectories clustering \cite{li2006coarse, wang2020clustering,martinsson2018clustering,HoseiniRC21}. Some  methods use Hidden Markov Models (HMM) to  deal with the sequential aspects of time series and trajectories, which are usually computationally expensive \cite{liu2019driving,takano2008recognition,Wang_2018}. Recent work uses Mixture of Hidden Markov Models (MHMM) that has shown promising results \cite{martinsson2018clustering}. An advantage of HMM is simplicity and  interpretation.

TimeNet, proposed in \cite{malhotra2017timenet}, is a multilayered recurrent neural network for feature extraction from time series. The authors demonstrate the performance of TimeNet  on tasks such as classification and clustering where they compute an embedding based on  t-SNE \cite{maaten2008visualizing}. Embedding the time-series has been also studied in \cite{nguyen2017m} where the proposed method, called m-TSNE, uses Dynamic Time Wrapping (DTW) \cite{DTW2018} as a metric between multidimensional time series embedded by t-SNE. The work in \cite{HoseiniRC21} develops a trajectory clustering method based on embedding temporal relations via DTW  and deep learning, and then extracting the transitive relations via minimax distances \cite{Chehreghani17AAAI,Chehreghani20MLJ}.
Finally, it is notable that clustering sequential data clustering is beyond trajectory analysis and has been studied for example for tree-structured sequences in \cite{ChehreghaniRLC07}.

\begin{figure}[th!]
     \centering
     \subfigure[Longitudinal distance w.r.t. time]{\includegraphics[width=0.4\textwidth]{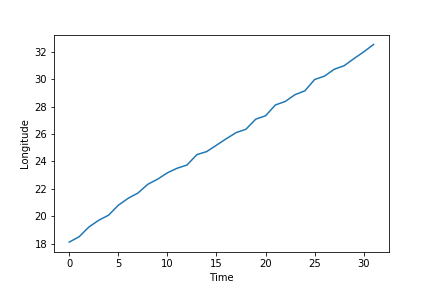}}
     \subfigure[Lateral distance w.r.t. time]{\includegraphics[width=0.4\textwidth]{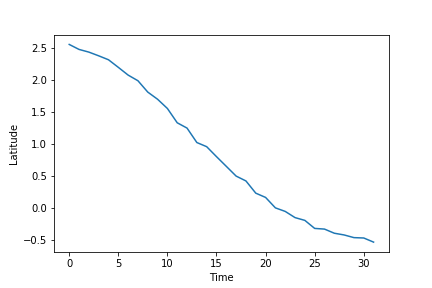}}
     \subfigure[Lateral distance w.r.t. longitudinal distance]{\includegraphics[width=0.4\textwidth]{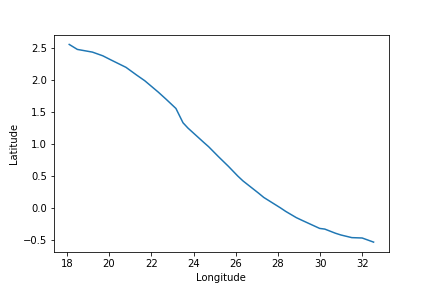}\label{fig:traj-example:latlog}}
     \subfigure[Lateral distance w.r.t. longitudinal distance]{\includegraphics[width=0.4\textwidth]{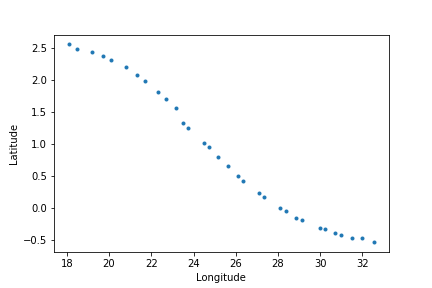}\label{fig:traj-example:latlogt}}
\caption{The extracted cut-in in different forms.}
\label{fig:traj-example}
\end{figure}

\section{Driving Scenario Data Source}
\subsection{Scenario Extraction}
The objects' trajectories are extracted from the raw data (sensor measurements) and fused sensor data, which are mounted on the ego car. Thus, the reference/coordinate system is the ego car. Since it is a moving reference, then all the measured signals (i.e., the position of the surrounding cars/objects) are relative with respect to the ego car. These trajectories can vary in length from 1 second up to 1 hour. The length depends on how long the object is tracked by the ego-vehicle in the field of view (FoV). The specific scenarios of high interest are cut-ins. There are many different definitions of what constitutes a cut-in. We define them as vehicles that approach the ego vehicle from the left lane and then overtake the ego vehicle by switching to its lane. Therefore, the cut-ins vary in aggressiveness. More specifically, our definition of a cut-in also requires the vehicle to stay in front of the ego vehicle for at least 2 seconds. An example of the extracted trajectory is illustrated in Fig. \ref{fig:traj-example}. Note that plotting trajectories as a line as shown in Fig. \ref{fig:traj-example:latlog} has the disadvantage of \emph{eliminating} the time component as compared to Fig. \ref{fig:traj-example:latlogt}. However, we find this way more expressive, as otherwise, it becomes extremely cluttered  when multiple trajectories overlap.

\subsection{On the Issue of Variable-Length Trajectories}
One of the main issues in analyzing the trajectories is the variable-length input/output, which in our case varies from 30 to 70 time frames (from 3 seconds to 7 seconds given the sampling rate of 10Hz). One solution is to train the model with padding. To apply padding, a pad token has to be defined. For instance, in natural language processing, it is common to employ word embedding and then to use zero vectors as a pad token \cite{cho2014learning}.
Unfortunately, it is not a trivial task to define a pad token in case of real coordinates as any pair of real numbers is a realistic point in space. A possible solution is to pad sequences with the last point. However, it does not seem a feasible approach in our case due to the high variation in length (the shortest sequence after padding will contain more than 50$\%$  pad tokens). These paddings not only may affect the distribution of the generated samples significantly, but also might call for post-processing of the samples. For example, if the last $n$ points are the same they should be considered as padding and erased. This yields an intrinsic problem as the definition of `being the same' is non-obvious in particular when some noise is added during the generation. Such problems can be avoided by feeding the sequences to the model one-by-one. However, this approach will greatly decline the performance.

Thus, in the proposed approach, we group the sequences with the same length together to form a batch. In this way, we train a model for the whole data but with different batches, where each batch represents a specific trajectory length. For example, assume trajectories of the following lengths (1-30 denotes that the trajectory number 1 has length 30): 1-30, 2-32, 3-32, 4-32, 5-34, 6-34, 7-34, and 8-34. Then, the following batches are formed: $\{1\},
 \{2,3,4\}, \{5,6,7,8\}$. If one propagation through the model takes 1 unit of time, then training the model with these batches would take 3 units of time, compared to 8 units of time with the one-by-one trajectory processing approach. The next steps depend on the architecture of the generative model, to be studied later.

\section{Trajectory Generation Framework}
In order to model and generate scenarios of trajectories with different lengths, we develop and propose two methods: i) An architecture based on combined Recurrent Autoencoder with GANs, where to obviate the variable length issue, the GAN is trained to learn/generate the hidden representation of original trajectories, instead of the original sequential data. ii) A Recurrent Conditional GANs (RC-GAN) architecture that enables us to generate driving sequences with pre-specified length, which is a desirable and useful feature when generating test cases for AD verification.
In the following, we explain on each of the two  methods in detail.

\subsection{The Architecture: Autoencoder with GANs (AE-GAN)}
\begin{figure*}[thb!]
    \centering
    \includegraphics[width= 0.9\textwidth]{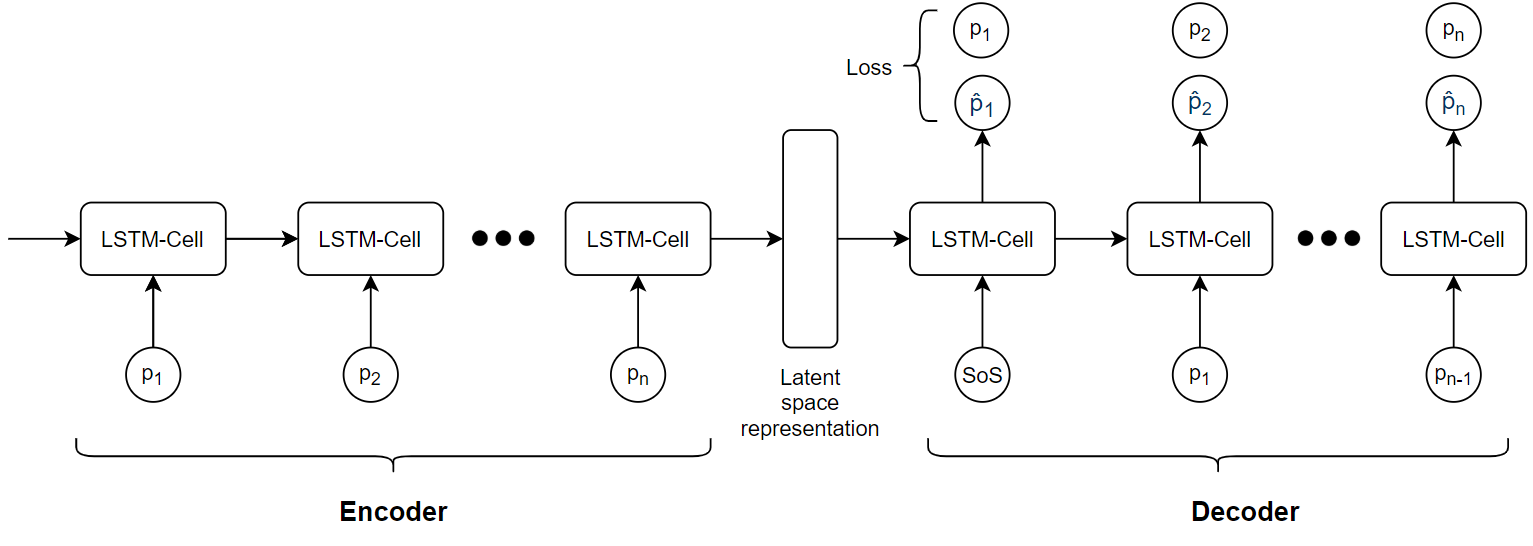}
    \caption{Schematic structure of recurrent autoencoder.}
    \label{ae-structure}
\end{figure*}

This solution is based on the architecture proposed for text generation in \cite{donahue2018adversarial}. It consists of an autoencoder for time series as shown in Fig. \ref{ae-structure} and GAN for latent space representation and data generation.

We adapt and extend this architecture to deal with variable-length input/output. It is essential to know the length of the sequence in order to bring it back from the latent-space representation. During the autoencoder training, the length is known from the input, but for the artificial latent-space vectors generated by GANs it is necessary to estimate the length of the trajectory. We address this issue by training a separate feed-forward neural network to estimate the lengths based on the latent space representation.

\begin{figure}[thb!]
    \centering
    \includegraphics[width=0.55\textwidth]{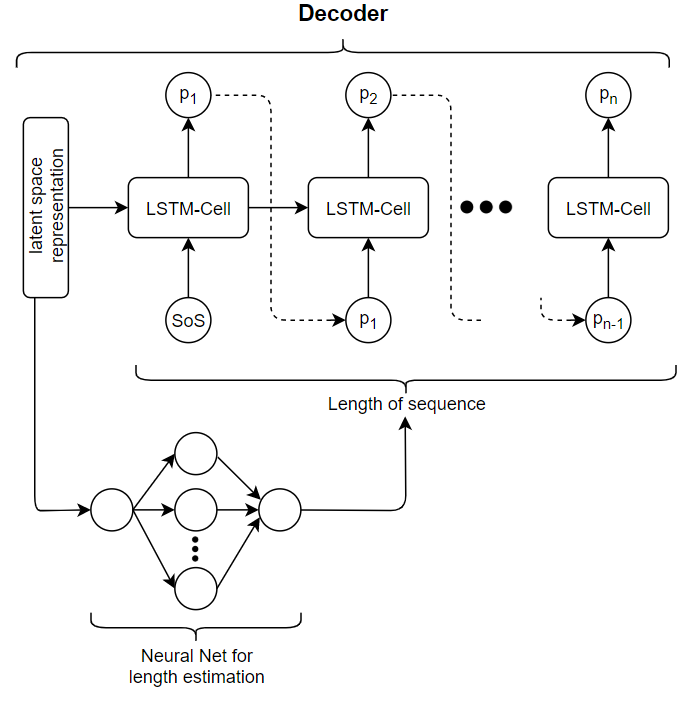}
    \caption{Structure of the decoder part of the autoencoder combined with a Length Estimator to reconstruct trajectories from latent space representation.}
    \label{v-to-t}
\end{figure}

Hence, once the autoencoder is trained, all trajectories are encoded to the latent space using the encoder. During this process, the length of each trajectory is stored. Thus, two sets are created: $X$: the set of latent representations, and $Y$: the set of the lengths for each trajectory. With these sets, the task of length estimation from latent space can be considered a supervised regression task which can be solved using a feed-forward neural network, as shown in Fig. \ref{v-to-t}.

At this stage, GANs are used to generate new latent-space representations. Even though it seems reasonable to implement both the generator and the discriminator as standard fully-connected neural networks, we will also investigate the ResNet model \cite{he2016deep} to mitigate the problems related to gradient instability, similar to the work in \cite{donahue2018adversarial}. To train GANs, we consider two alternatives: the standard GAN and the Wasserstein GAN with Gradient Penalty (WGAN-GP) \cite{GulrajaniAADC17}.

As the latent-space representation is generated, it is used as an input to the decoder. To determine how many times to apply LSTM cell, we employ the aforementioned neural network in Fig. \ref{v-to-t}.

\subsection{The Architecture: Recurrent Conditional GANs (RC-GAN)}
\begin{figure}[thb]
    \centering
    \includegraphics[width=0.65\textwidth]{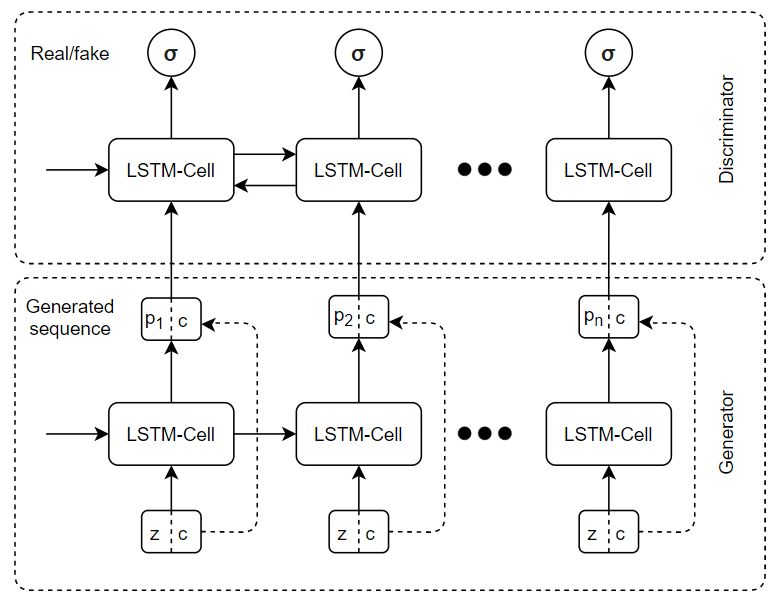}
    \caption{The schematic structure of RC-GAN.}
    \label{rec-gan}
\end{figure}

Recurrent GANs (RecGANs) have shown promising results for generating time series in several applications such as music generation \cite{mogren2016c}, real-valued medical data \cite{esteban2017real} and sensor error modeling \cite{arnelid2019recurrent}. In this paper, we adapt them for our task as follows. Both the generator and discriminator are Recurrent NNs (RNNs) based on LSTM cells. For the discriminator, we choose a bidirectional architecture. At every time step $i$, each LSTM cell in the generator receives a random input (i.e., $z$ drawn from $\mathcal{N}(0,1)$) and the hidden vector from the previous cell, in order to generate $p_i$. For the first cell, the previous hidden vector is initialized to $0$. The sequence $p_1...p_n$ forms the final trajectory, that is passed to the discriminator. Then, the discriminator computes a sequence of probabilities ($\sigma$ in Fig. \ref{rec-gan}) identifying whether the trajectory is real or fake. The ground truth is a sequence of ones for the real trajectories and zeros for the fakes. RecGANs can also be conditional. With Recurrent Conditional GANs (RCGANs), the condition might be passed as an input into each cell  for both the discriminator and the generator. As shown in \cite{esteban2017real}, the condition can be simply concatenated to the generator's input and output (Fig. \ref{rec-gan}). This allows the discriminator to distinguish between reals and fakes w.r.t. to condition, that in turn forces the generator to produce samples w.r.t condition. To adapt this architecture to our task,  we use the length of a trajectory as a condition and attach it to the input.

\subsection{Evaluation}
An important and challenging task is to choose proper metrics to evaluate the quality of the generated trajectories \cite{lucic2018gans}. One might first evaluate the results via visualization to see whether they do make sense. However, as the results improve it becomes harder to determine precisely how good they are. Thus, it is important to consider quantitative evaluation metrics as well, such that one can objectively quantify the similarity of the generated trajectories with the original ones.

One commonly used method to measure similarities between time series is Dynamic Time Warping (DTW).
Thus, to compare sets of time series, we build a matrix of pairwise DTW distances between the samples from the two sets as shown in Table \ref{Table:pairwisedist}. Such a matrix can be used to find the most similar samples from the two investigated sets. In addition, we analyze this pairwise matrix using the following two metrics. In the following, we describe two methods for the analysis of results based on matching the time series.

\begin{table}[b]
\caption{Pairwise distances between trajectories from two sets, here $tr$ stands for the trajectory.}
\begin{center}
\begin{tabular}{|c|c|c|c|c|c|}
\cline{3-6}
\multicolumn{2}{c|}{} & \multicolumn{4}{c|}{Set 1} \\
\cline{3-6}
\multicolumn{2}{c|}{} & $tr_1$ & $tr_2$ & \dots & $tr_n$ \\
\hline
\multirow{4}{*}{\rotatebox[origin=c]{90}{Set 2}} & $tr'_1$ & $dist(tr'_1, tr_1)$ & $dist(tr'_1, tr_2)$ & \dots & $dist(tr'_1, tr_n)$ \\
\cline{2-6}
& $tr'_2$ & $dist(tr'_2, tr_1)$ & $dist(tr'_2, tr_2) $ &\dots & $dist(tr'_2, tr_n)$ \\
\cline{2-6}
& \dots &  \dots &  \dots & $ \ddots$ &  \dots  \\
\cline{2-6}
& $tr'_n$ & $dist(tr'_n, tr_1)$ & $dist(tr'_n, tr_2) $ &\dots & $dist(tr'_n, tr_n)$ \\
\hline
\end{tabular}
\end{center}
\label{Table:pairwisedist}
\end{table}

\subsubsection{Matching + Coverage}
We match each sample from the generated set (called $GS_M$ with $M$ samples) with the closest sample from the real set (called $RS_N$ with $N$ samples).
The matching criterion is defined by

\begin{equation}
matching =\frac{\sum_{i}^{M}{\min_{j}{ (dist(GS_i, RS_j)) }}}{M}
\end{equation}

Even if reasonable results are achieved with this metric, it does not necessarily indicate that the model performs well, since many generated samples can be `mapped' to the same real sample. In this case, the coverage of the model is low. Thus, we also measure a coverage metric as follows.

\begin{equation}
coverage = \frac{|argmin_j(dist(GS_i, RS_j)), \forall i = \overline{1,M}|}{N}
\end{equation}

However, even the combination of these two metrics still has shortcomings. For instance, if there are two (or more) similar samples in the real set and many generated samples are `mapped' to one of the real samples, then the coverage decreases. However, this does not mean that the model performs poorly. Since the sets are very diverse, we consider $GS_M$ and $RS_N$ with $M>N$. In our experiments, we use $M=4*N$

\subsubsection{One-to-One Matching with Hungarian Method}
The Hungarian algorithm is a matching method for one-to-one matching. Applying it to Table \ref{Table:pairwisedist} yields mapping each sample from the generated set to exactly one sample from the real set. This mapping ensures that the sum of distances of the paired samples is minimal. The main disadvantage of this method is the sample distributions in the real and generated sets may not be identical. For example, the last 10\% of the matched samples can be outliers that are from irrelevant parts of the distribution.

Once the aforementioned metrics are defined it is still an open question which ground truth should be used as a reference to compare the results with. To address this question we spilt the real dataset into different subsets and apply these metrics among the different subsets. We use the results as a baseline when analyzing the trajectories obtained from the generation models.

\section{Exploratory Analysis of Latent Space Representations}
GANs provide an unsupervised learning approach to generate samples consistent with a given set of real trajectories.
In the following, we investigate other unsupervised learning methods in the context of clustering and outlier detection, in order to obtain exploratory insights from the driving trajectories of the surrounding objects, collected by sensors of the ego car. In addition, as mentioned before, these methods can be useful for the safety verification of AD.
Consistent with the proposed GAN architectures, our clustering and outlier analysis mechanisms are also performed based on the latest space representation obtained from training the autoencoder. This helps us to benefit from the representation that encodes the temporal aspects of the trajectories and simplifies the process. We note that the solution from the latent space representations can be transferred to the original trajectories, in order to provide a solution in the original data space.

Performing for example the clustering on original trajectories might require methods such as DTW to model the temporal aspects first. Then, we may apply a  high-dimensional data visualization and grouping method such as t-SNE \cite{maaten2008visualizing}. The method proposed in \cite{nguyen2017m} and called m-TSNE applies t-SNE to DTW-based (dis)similarity matrix. However, it is computationally very expensive, when working with a large number of scenario trajectories. In the case of $n$ trajectories of length $m$, it needs to calculate $n^2$ pairwise distances and each distance is computed with DTW that runs in $O(m^2)$. Thus, the overall performance is $O(n^2m^2)$. In our setup, we have relatively long trajectories (50 time step on average) and relatively a large number of them. Therefore, we employ the already trained autoencoder which encodes the temporal dependencies properly. We assume that clustering and outlier detection at latent space is an easier task with reasonable computational costs.

\subsection{Clustering on Latent Space with Autoencoder}

\begin{figure}[bth]
  \begin{center}
    \includegraphics[width=0.9\textwidth]{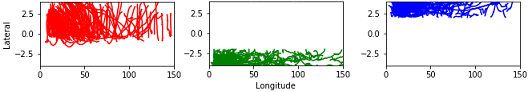}
  \caption{Three types of explicit-rule extracted trajectories: cut-ins,
  right- and left drive-by.}
  \label{fig:labeledtrajs}
  \end{center}
\end{figure}

Here, we study clustering the latent space representation of trajectories, obtained by recurrent autoencoder. For this purpose, we extract three types of scenarios: cut-in,
  left drive-by and right drive-by, as shown in Fig. \ref{fig:labeledtrajs}.

\begin{figure}[t]
    \centering
    \includegraphics[width=0.75\textwidth]{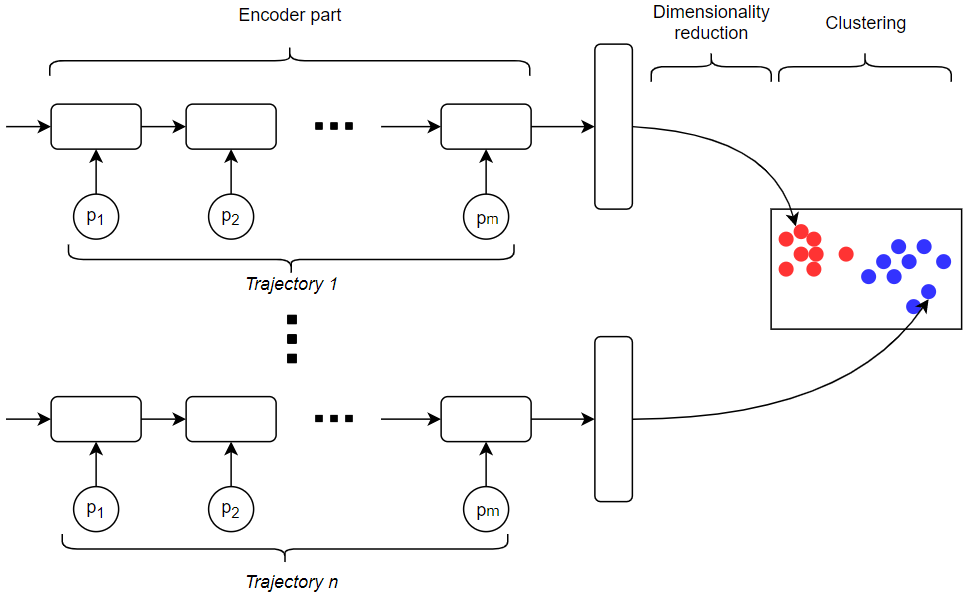}
    \caption{Clustering process with AE}
    \label{AE-clustering-scheme}
\end{figure}

Drive-by occurs much more frequently. Thus, the cut-in set contains fewer trajectories compared to the drive-by set. To address this issue, we do oversampling for the cut-in set and undersampling for the drive-by sets. Then, we train the autoencoder and encode all the trajectories. This step converts all trajectories to fixed-size vectors. Even though the resultant latent space representations are fixed-size, they are still high dimensional. Thus, it might be challenging to cluster them directly with distance-based algorithms such as K-means or DBSCAN. Therefore, we reduce the dimensionality with methods such as Principal Component Analysis (PCA) and t-SNE. Finally, we apply the clustering method (e.g., DBSCAN\cite{ester1996density}) and analyze the results. The procedure is described in Fig. \ref{AE-clustering-scheme}, where the encoder part is the same for all trajectories and $p_i$ represents lateral and longitudinal positions in our case. In this way, each trajectory is mapped to a two-dimensional representation.

\subsection{Outlier Detection with Autoencoder}
Being able to detect anomaly driving patterns and outliers among scenario trajectories is very valuable in different aspects in particular when the data is imbalanced. First, it can be used to assess the quality of the original data and to find possible sensor reading anomalies/errors. A more detailed investigation can be then performed afterwards, by checking the camera videos and LIDAR sensor reads, in order to gain more insights about the detected anomalies. Second, it can be used to find the minority sub-groups in the data. For example, aggressive driving of the surrounding vehicles w.r.t. ego vehicle is an important test case for verification of AD functionality. Having this information about anomalies can also improve the quality of the generation process by treating them differently. Third, it is often important to determine if a set of data is homogeneous and balanced before any statistical technique is applied. Finally, this information can help us to re-calibrate our explicit rule-based cut-in finder functions. These functions, which usually are defined based on a hard threshold for a  scenario, might perform poorly on anomalies.

In the following, we describe a method to detect and analyze outliers using the trained autoencoder.
We assume a high reconstruction loss in the autoencoder implies some anomaly in the respective trajectory, i.e., the sample is an outlier. We may define a threshold for the loss and consider all the trajectories that yield a higher reconstruction loss than the threshold as outliers. However, choosing a hard threshed might be nontrivial. On the other, a hard assignment might not be very robust. Therefore, we follow a `soft' approach instead, where we compute the probability of a trajectory ($s_i$) being an outlier.

\begin{equation}
    p(s_i \textit{ is outlier}) = \frac{\exp(l(s_i))}{\exp(l(s^*))}\;\; ,
\end{equation}

where $s^*$ corresponds to the trajectory with maximal reconstruction loss. We note that instead of normalization by $\exp(l(s^*))$, one may use any other normalization which might makes sense depending on the context.

\section{Experimental Results}
In this section, we investigate the different aspects of the proposed methods on real-world data and scenarios.

Fig. \ref{fig:ae-demo:real} shows 100 real trajectories wherein a cut-in occurs. It is clear that the distribution is not even and uniform. There are a lot more samples in the 20-60 meters longitudinal region while only a few are seen past 100 meters. Another observation is that the majority of the trajectories have a trend to increase in the longitudinal distance through time. However, there are several samples for which the longitudinal distance decreases instead. This can be interpreted as the cut-ins wherein the tracked vehicle accelerates or decelerates respectively. It seems worth checking if the proposed models capture these different trends and outliers.

\subsection{Autoencoder}
\begin{figure}[tbh!]
  \begin{center}
  \subfigure[Real Trajectories]{\includegraphics[width=0.4\textwidth]{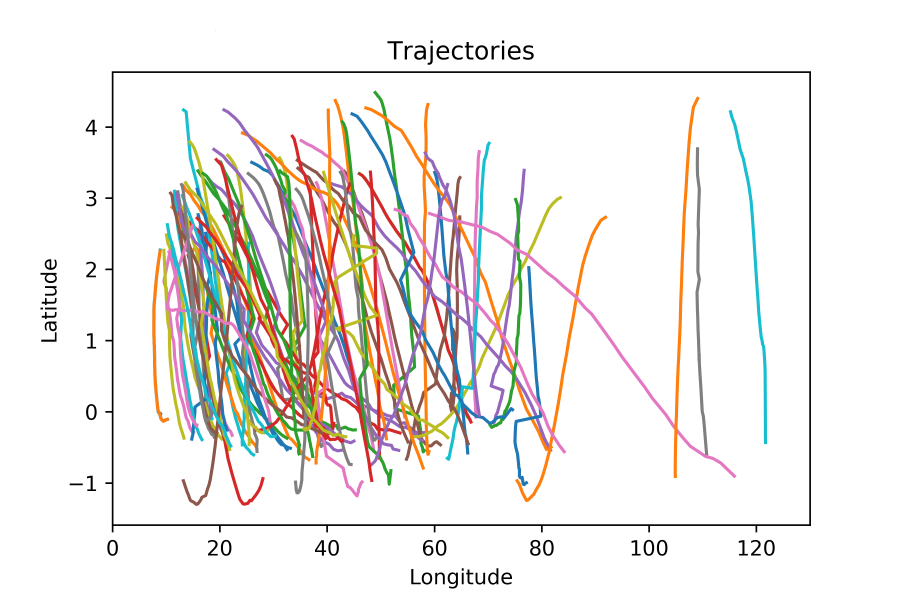}\label{fig:ae-demo:real}}
  \subfigure[Reconstructed Trajectories]{\includegraphics[width=0.4\textwidth]{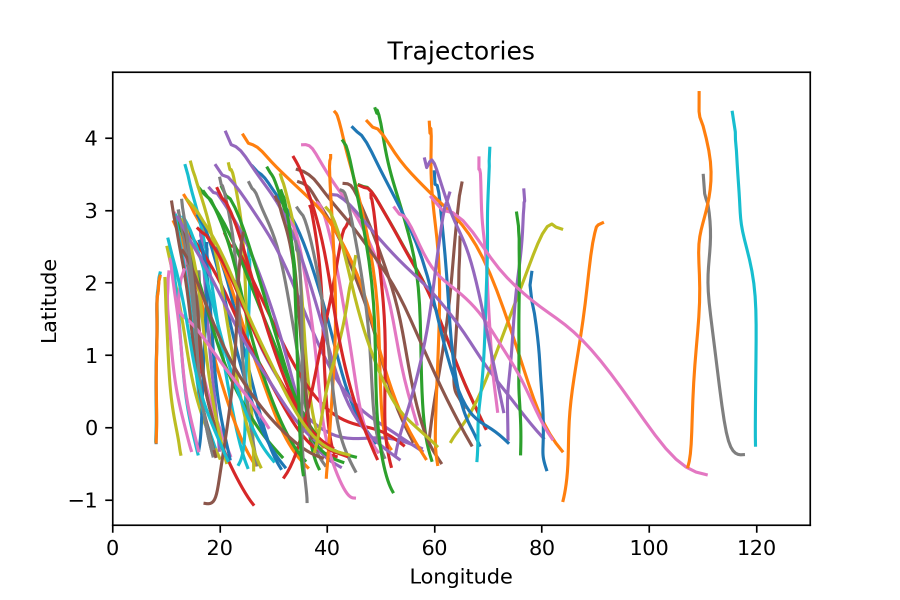}\label{fig:ae-demo:rec}}
  \caption{100 real trajectories (a) and their reconstruction by autoencoder (b).}
  \label{fig:ae-demo}
  \end{center}
\end{figure}

We start with examining the results from the autoencoder.  Fig. \ref{fig:ae-demo} illustrates the real and reconstructed trajectories. The main difference between them is the smoothness of the reconstructed ones which is a typical and expected property of an autoencoder.

We perform two experiments: the first experiment with the trajectories from 3 to 5 seconds, and the second experiment with the trajectories from 3 to 7 seconds. In both cases, a two-layer LSTM cell is used. The loss values with respect to different sizes of hidden states are shown in Table \ref{Table:ae-hiddensize}. For the first experiment, a hidden state of size 32 is sufficient and produces meaningful results from a visual inspection point of view. However, for the second experiment, we choose the size of the hidden state to be 64 as it decreases the loss drastically. Note that close loss values for trajectories with different lengths does not necessarily imply the same performance for the autoencoder, since the mean is calculated with respect to a different number of samples. 

\begin{table}[tbh]
\centering
\caption{Comparison between different sizes of hidden state (hs) for two sets consisting of trajectories between 3 to 5 seconds (I) and 3 to 7 seconds (II).}
\begin{tabular}{|l|l|l|l|l|} \hline
Size of hs & Val.Loss I & Train Loss I & Val.Loss II & Train Loss II \\ \hline
32  & 0.0633 & 0.0652 & 0.0648 & 0.0637 \\ \hline
64  & 0.0619 & 0.0620 & 0.0469 & 0.0469 \\ \hline
128 & 0.0601 & 0.0610 & 0.0451 & 0.0471 \\ \hline
\end{tabular}
\label{Table:ae-hiddensize}
\end{table}

\begin{figure}[tbh]
     \centering
     \subfigure[Length from 3 to 4 seconds]{\includegraphics[width=0.45\textwidth]{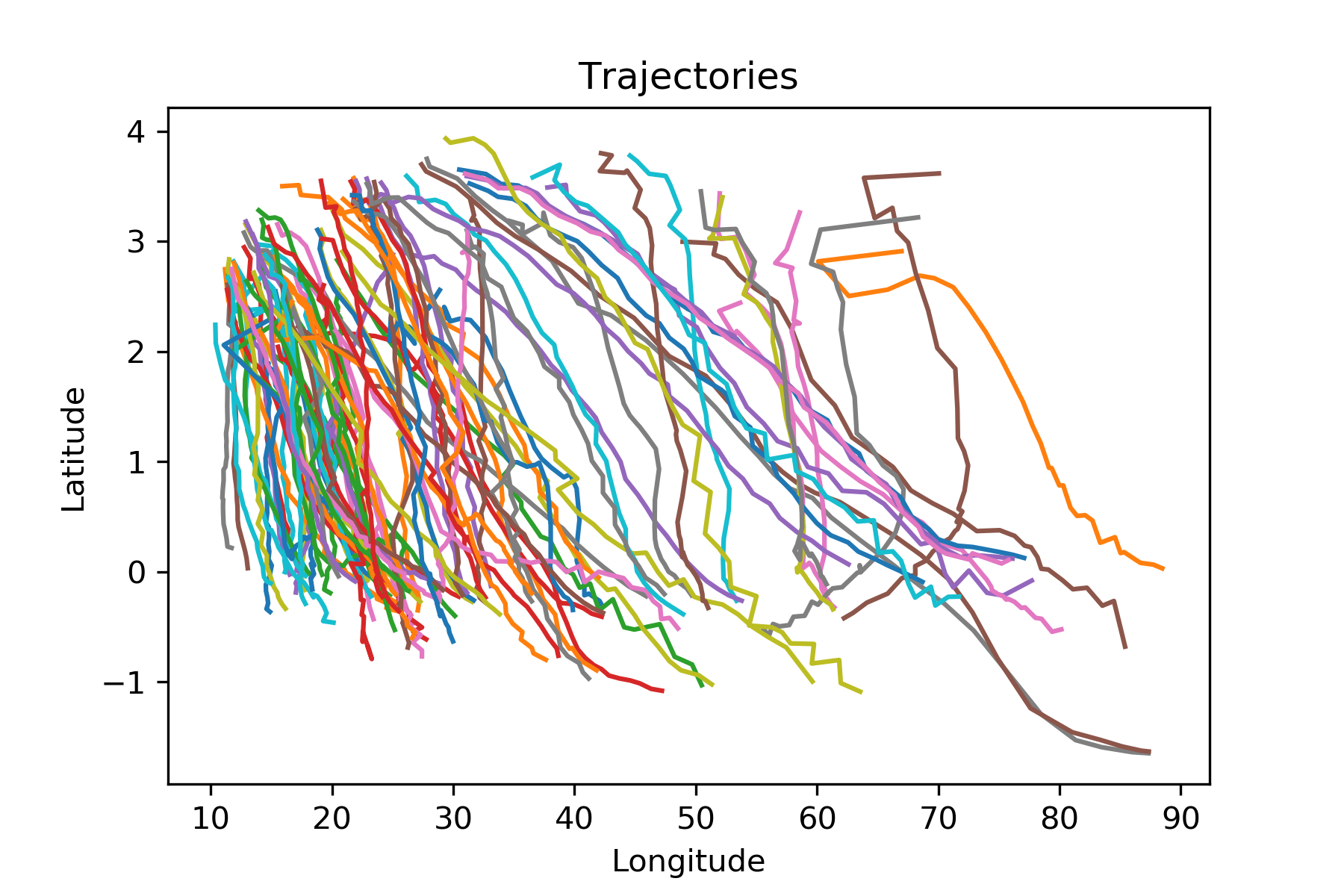}}
     \subfigure[Length from 6 to 7 seconds]{\includegraphics[width=0.45\textwidth]{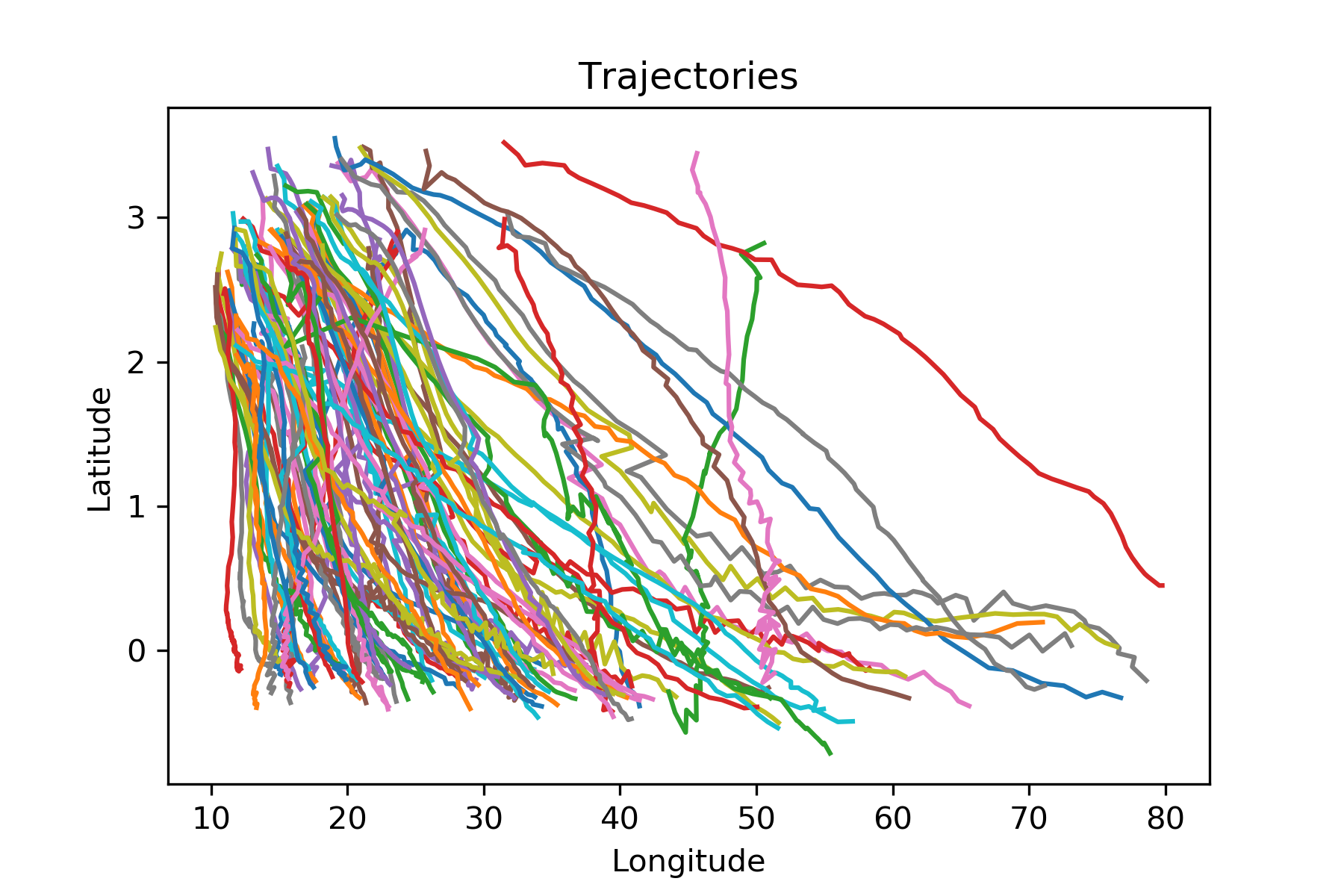}}
\caption{The trajectories generated with RC-GANs.}
\label{fig:recdemo}
\end{figure}

\subsection{Generative Models: AE-GAN \& RC-GAN}
\begin{figure}[thb]
     \centering
     \subfigure[Example of 100 trajectories from 3 to 5 seconds generated with AE-GAN]{\includegraphics[width=0.45\textwidth]{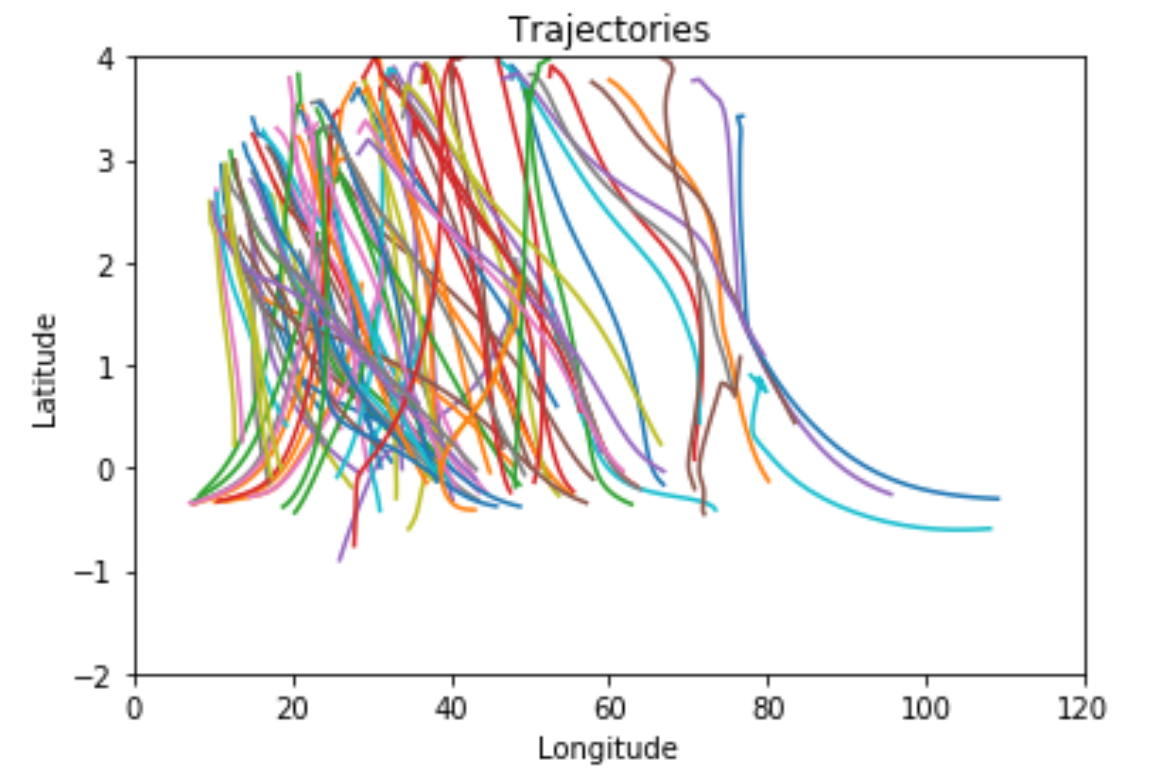}\label{fig:aeGAN:a}}
     \subfigure[Example of 100 trajectories from 3 to 7 seconds generated with AE-WGAN-GP]{\includegraphics[width=0.45\textwidth]{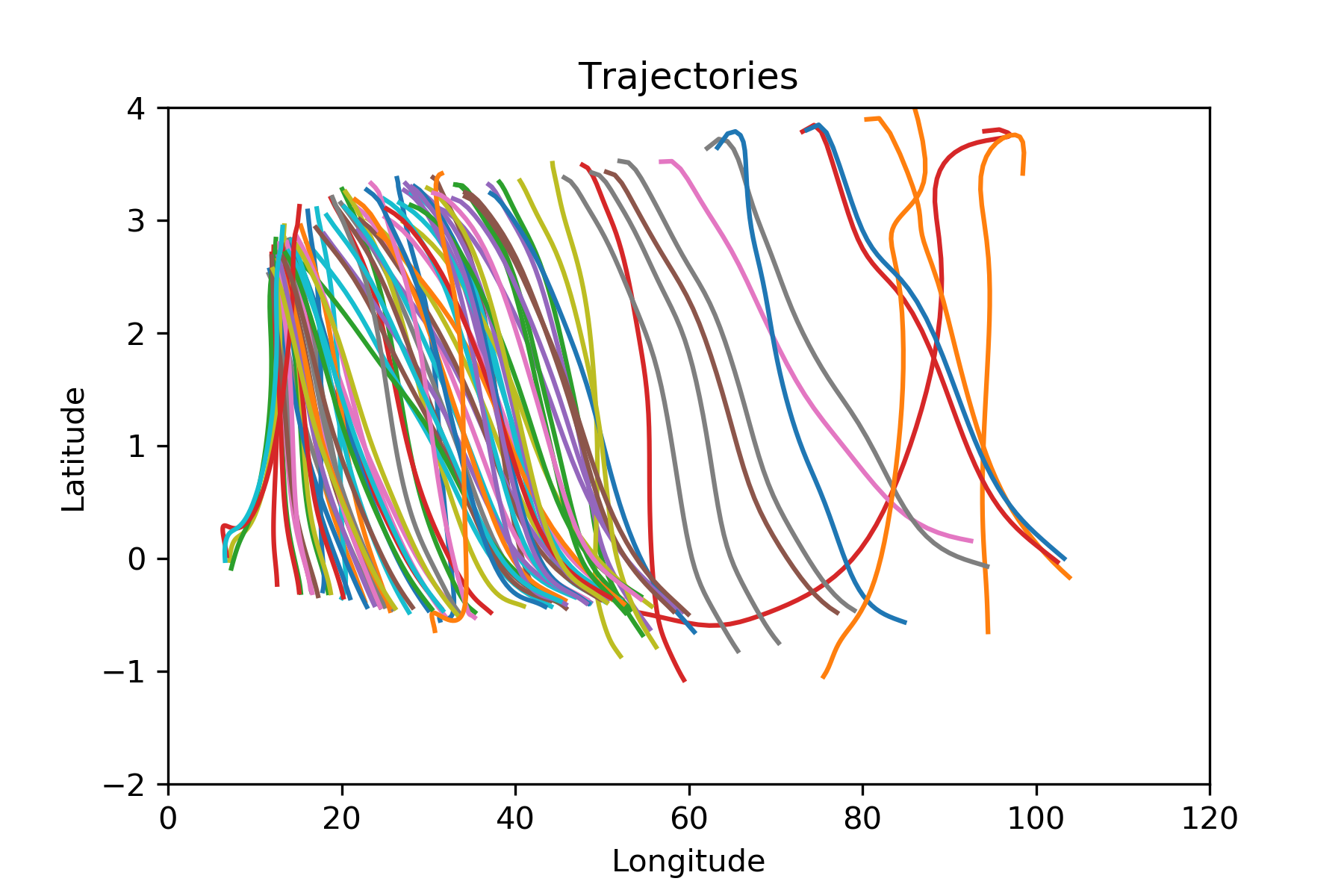}\label{fig:aeGAN:b}}
\caption{The trajectories generated with AE-GAN and AE-WGAN-GP.}
\label{fig:aeGAN}
\end{figure}

Fig. \ref{fig:recdemo} and Fig. \ref{fig:aeGAN} illustrate the trajectories generated respectively with RC-GAN and AE-GAN. Obviously, both models capture the trends of the data. However, some of the generated trajectories can be distinguished from the real ones. The samples from RC-GAN are noisier compared to the real ones, while samples generated with AE-GAN are more smooth. From a visual inspection, both models seem to capture the distribution of the trajectories. Similar to the real dataset, there are more trajectories generated close to the ego-vehicle and less further away. Both models also generate accelerating and decelerating cut-ins.

Our proposed RC-GAN is conditioned on the length of the trajectory. It is therefore possible to generate trajectories with a pre-specified length. This condition works as expected. As it can be seen in Fig. \ref{fig:recdemo}, all trajectories end up in region 0 (from a lateral perspective), which means they are complete cut-ins. For example, there are no trajectories that are just truncated halfway after 3 seconds.

\begin{figure}[thb]
     \centering
     \subfigure[Generated trajectories with RC-GAN.]{\includegraphics[width=0.4\textwidth]{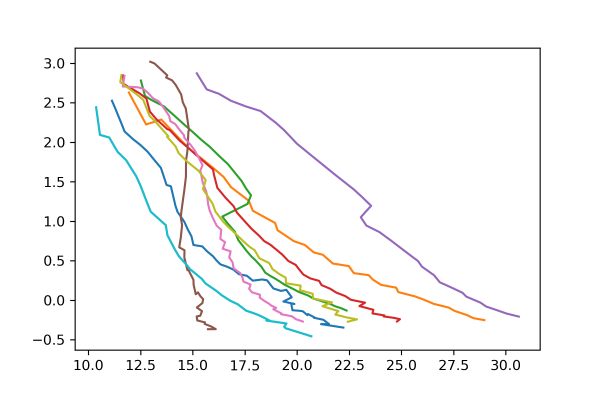}}
     \subfigure[Real trajectories closest to the trajectories presented in (a).]{\includegraphics[width=0.4\textwidth]{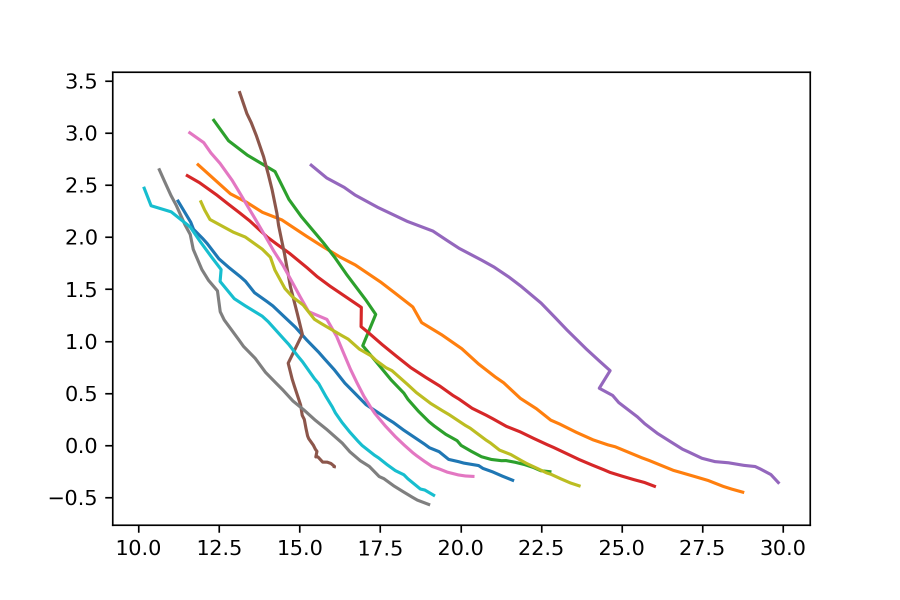}}
     \\
     \subfigure[Generated trajectories with AE-GAN.]{\includegraphics[width=0.4\textwidth]{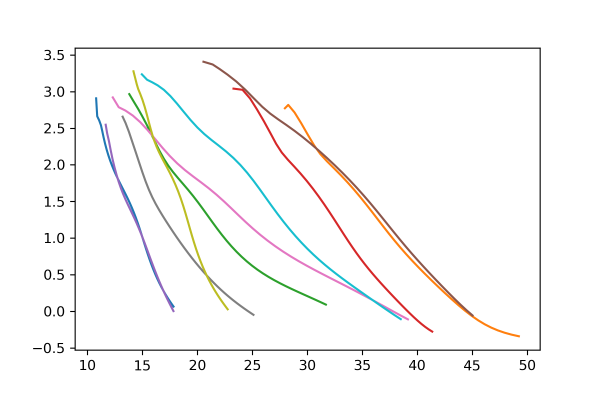}}
     \subfigure[Real trajectories closest to the trajectories presented in (c).]{\includegraphics[width=0.4\textwidth]{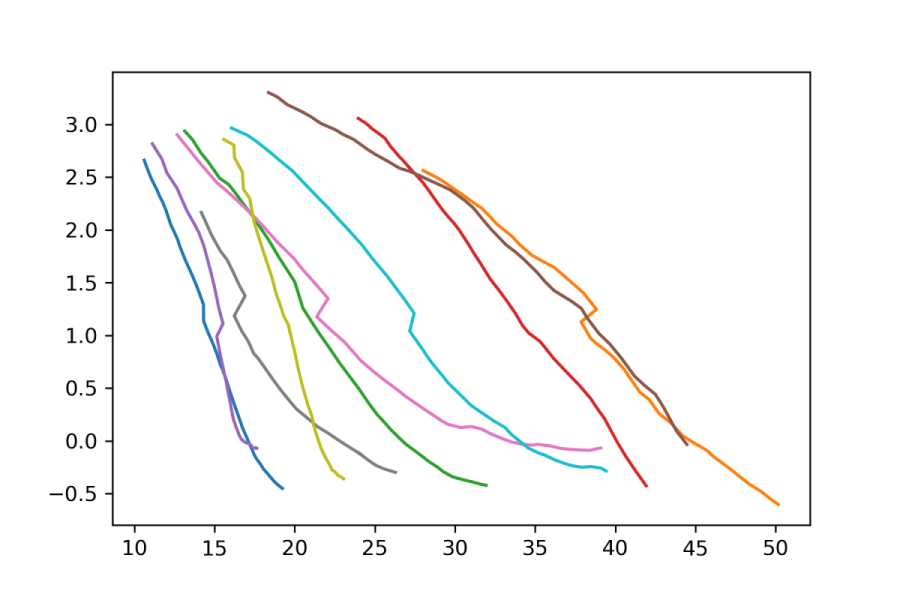}}
\caption{Illustration of 10 closest samples between generated and real sets.}
\label{fig:best10}
\end{figure}

With AE-GAN, we start our experiments for trajectories from 3 to 5 seconds with fully-connected neural networks and original GANs. The results of this experiment are illustrated in Fig. \ref{fig:aeGAN:a}. Unfortunately, this setting does not produce meaningful results for the 3 to 7 seconds trajectories. Thus, we experiment with the WGAN-GP and ResNet architecture for the generator and the discriminator. The ResNet architecture does not introduce any great improvement. However, WGAN-GP allows us to generate trajectories from 3 to 7 seconds as shown in Fig. \ref{fig:aeGAN:b}.
Fig. \ref{fig:best10} illustrates the trajectories with minimal DTW distances from RC-GAN and AE-GAN.

\subsection{Quantitative Comparisons}

We first apply the proposed metrics to the real set to obtain a baseline. Each experiment is done 5 times and the average score  together  with the maximum and minimum scores are reported in Tables \ref{table-res-M-C} and \ref{table-res-H} for the first and second metrics, respectively. Note that the experiments with RC-GAN and AE-WGAN-GP are performed for 3 to 7 seconds trajectories while the results with AE-GAN are only for 3 to 5 seconds trajectories.

Based on the results of the \emph{Matching+Coverage} metric in Table \ref{table-res-M-C}, RC-GAN yields the highest coverage and the lowest matching error among the generated sets. The matching metric for RC-GAN is even lower than the real set which can be explained by a lower coverage: calculating the average for about 60\% of the best matched samples in the real set produces a lower score. On the other hand, as discussed in Fig. \ref{fig:recdemo} and Fig. \ref{fig:aeGAN}, the AE-WGAN-GP trajectories are significantly smoother compared to the RC-GAN trajectories. Since the performance of AE-WGAN-GP is still close to the real sets in particular w.r.t. the \emph{Matching} metric, thus it might be preferred in practice.

\begin{table}[b]
\centering
\caption{The results for Matching and Coverage metric. Min, Max and Average are computed from  5 experiments.}
\begin{tabular}{|l||l|l|l|l|l|l|} \hline
Set      & \multicolumn{3}{c|}{Matching}  & \multicolumn{3}{c|}{Coverage} \\ \hline
         & Min   & Max   & Average & Min  & Max  & Average \\ \hline
Real Set (Baseline) & 39.78 & 45.46 & 43.28   & 0.85 & 0.89 & 0.875  \\ \hline
RC-GAN    & 29.08 & 31.93 & 30.37 & 0.63 & 0.68 & 0.66  \\ \hline
AE-GAN   & 48.85 & 56.13 & 53.31 & 0.39 & 0.45 & 0.42 \\ \hline
AE-WGAN-GP & 39.04 & 52.65 & 44.70 & 0.54 & 0.59 & 0.56  \\ \hline
\end{tabular}
\label{table-res-M-C}
\end{table}

\begin{table}[b]
\centering
\caption{The results for Hungarian distance. Min, Max and Average are computed from  5 experiments.}
\begin{tabular}{|l||l|l|l|l|l|l|} \hline
Set        & \multicolumn{3}{c|}{Hungarian} & \multicolumn{3}{c|}{ Hungarian (75\%)}\\ \hline
           & Min    & Max    & Average & Min    & Max    & Average \\ \hline
Real Set
(Baseline) & 62.01  & 84.58  & 75.2   & 29.84  & 40.14  & 36.10 \\ \hline
RC-GAN      & 330.38 & 500.66 & 430.33 & 121.14 & 151.83 & 138.56 \\ \hline
AE-GAN     & 330.3  & 424.59 & 358.69 & 121.94 & 141.74 & 130.02\\ \hline
AE-WGAN-GP & 159.15 & 284.36 & 229.91 & 63.47  & 102.68 & 79.32  \\ \hline
\end{tabular}
\label{table-res-H}
\end{table}

\begin{figure}[tbh]
  \begin{center}
    \includegraphics[width=0.85\textwidth]{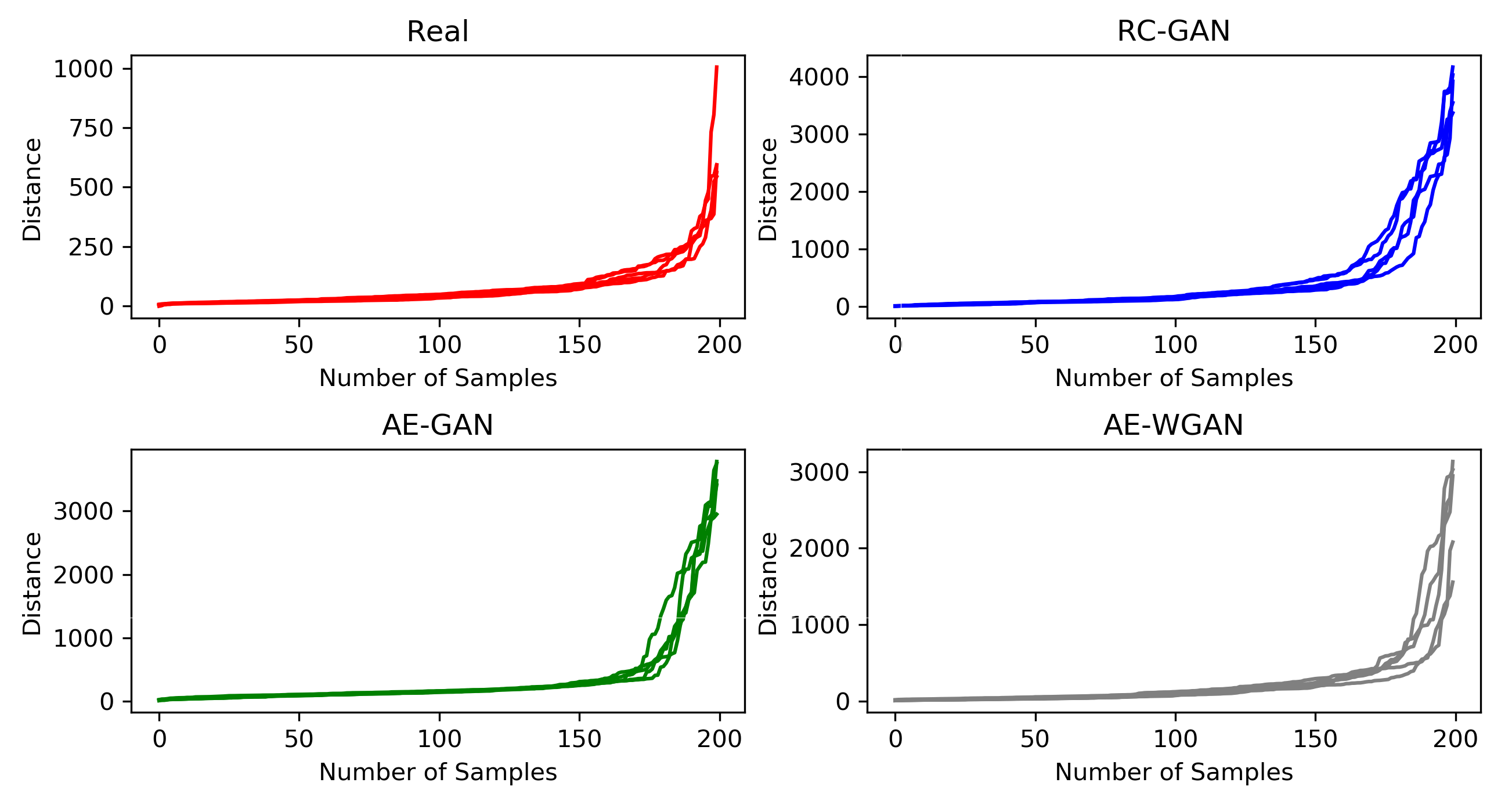}
  \caption{The pairwise distances between matched samples in one-to-one matching: real: Red, RC-GAN: blue, AE-GAN: green, AE-WGAN: grey.}
  \label{fig:treshold}
  \end{center}
\end{figure}

The results for the second metric that is based on one-to-one matching (i.e., the Hungarian method) are shown in Table \ref{table-res-H}. The best result belongs to AE-WGAN-GP which is 229.91. It is about three times higher than the result of the real set.  However, such behavior is expected due to an uneven distribution of trajectories. To obtain  more useful insights from the one-to-one matching, the distances between the matched samples are plotted in Fig. \ref{fig:treshold}. We can clearly see similar behavior between the generated and real sets. The scale of these graphs is different, and the matched distances for generated sets explode more compared to the real set. We assume this behavior is a combination of two factors: the dissimilar distribution of trajectories and some generated samples that are far from being realistic. From Fig. \ref{fig:treshold} it is observed that none of the plots explodes until 150 out of 200 samples. Thus, we also compute an average of the matched distances for only the first 75\% of samples, as shown in Table \ref{table-res-H}. According to these results, the generated samples (in particular by AE-WGAN-GP) are more consistent with the real trajectories.

\subsection{Clustering}

\begin{figure}[thb]
     \centering
     \subfigure[PCA]{\includegraphics[width=0.45\textwidth]{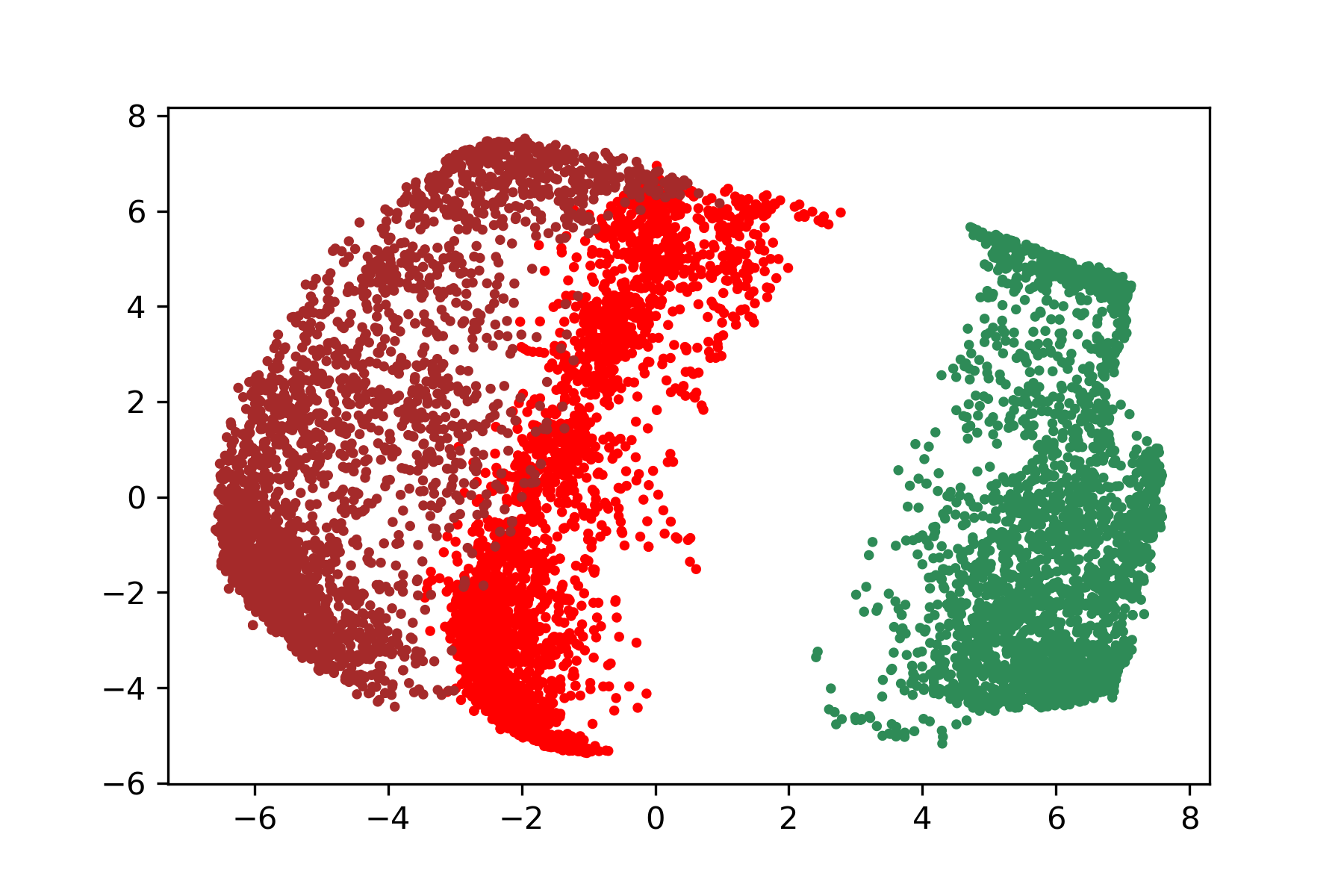}}
     \subfigure[SVD]{\includegraphics[width=0.45\textwidth]{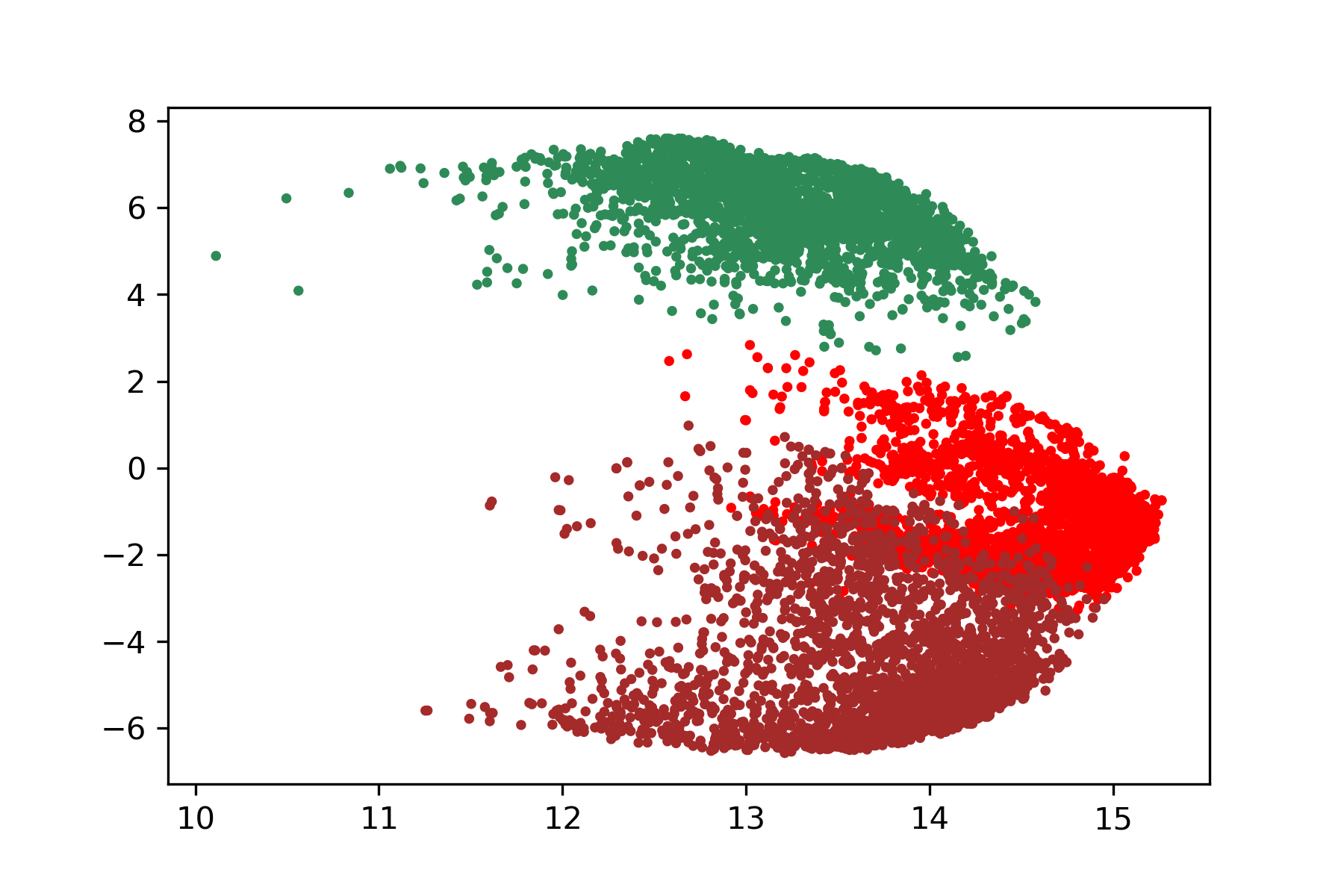}}
     \subfigure[t-SNE]{\includegraphics[width=0.45\textwidth]{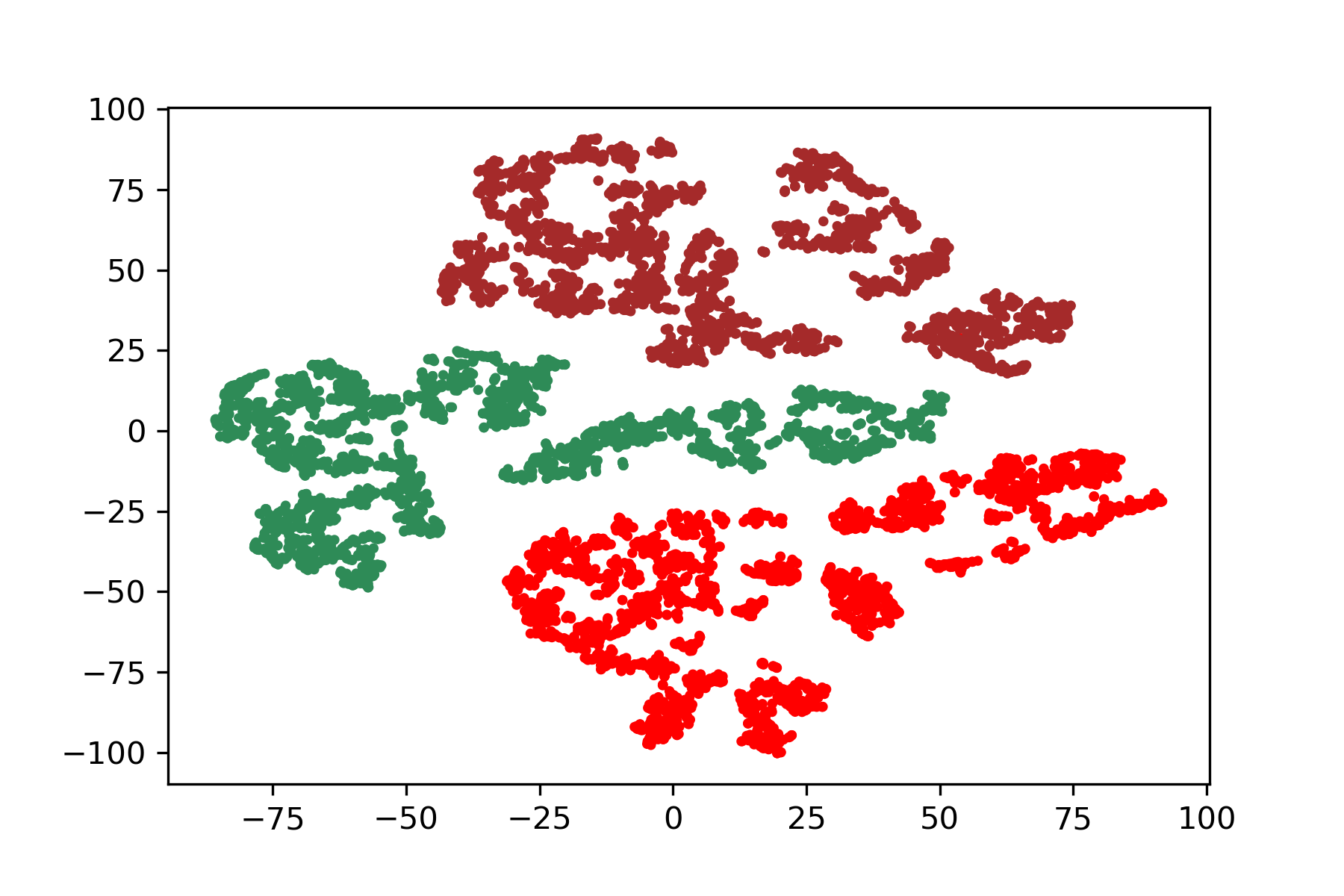}}
\caption{Latent space representation of trajectories in two dimensions with ground-truth labels. green: right drive by, purple: left, red: cut-in}
\label{fig:dimred}
\end{figure}

In the following, we investigate clustering and in particular the DBSCAN method on latent space representations. As mentioned before, to handle the high dimensionality issues, DBSCAN \cite{ester1996density} is used in combination with  dimensionality reduction techniques to reduce the number of dimensions: PCA, SVD and t-SNE. 
Fig. \ref{fig:dimred} shows the results of different methods in two dimensions. We observe that neither PCA nor SVD transform the data such that it can be clustered properly, i.e., the clusters have overlaps. PCA performs slightly better than SVD, thus we skip SVD.
Unlike PCA and SVD, t-SNE provides a non-overlapping and well-separated embedding.
With PCA and SVD, we obtain a diagonal matrix $\Sigma$ with singular values and based on them, it is possible to calculate a percentage of variance introduced by each component. This information can help to find an optimal number of principal components to capture enough information from the original data to distinguish clusters while at the same time avoid the curse of dimensionality.

\begin{figure}[tb]
     \centering
     \subfigure[DBSCAN applied to t-SNE embedding.]{\includegraphics[width=0.45\textwidth]{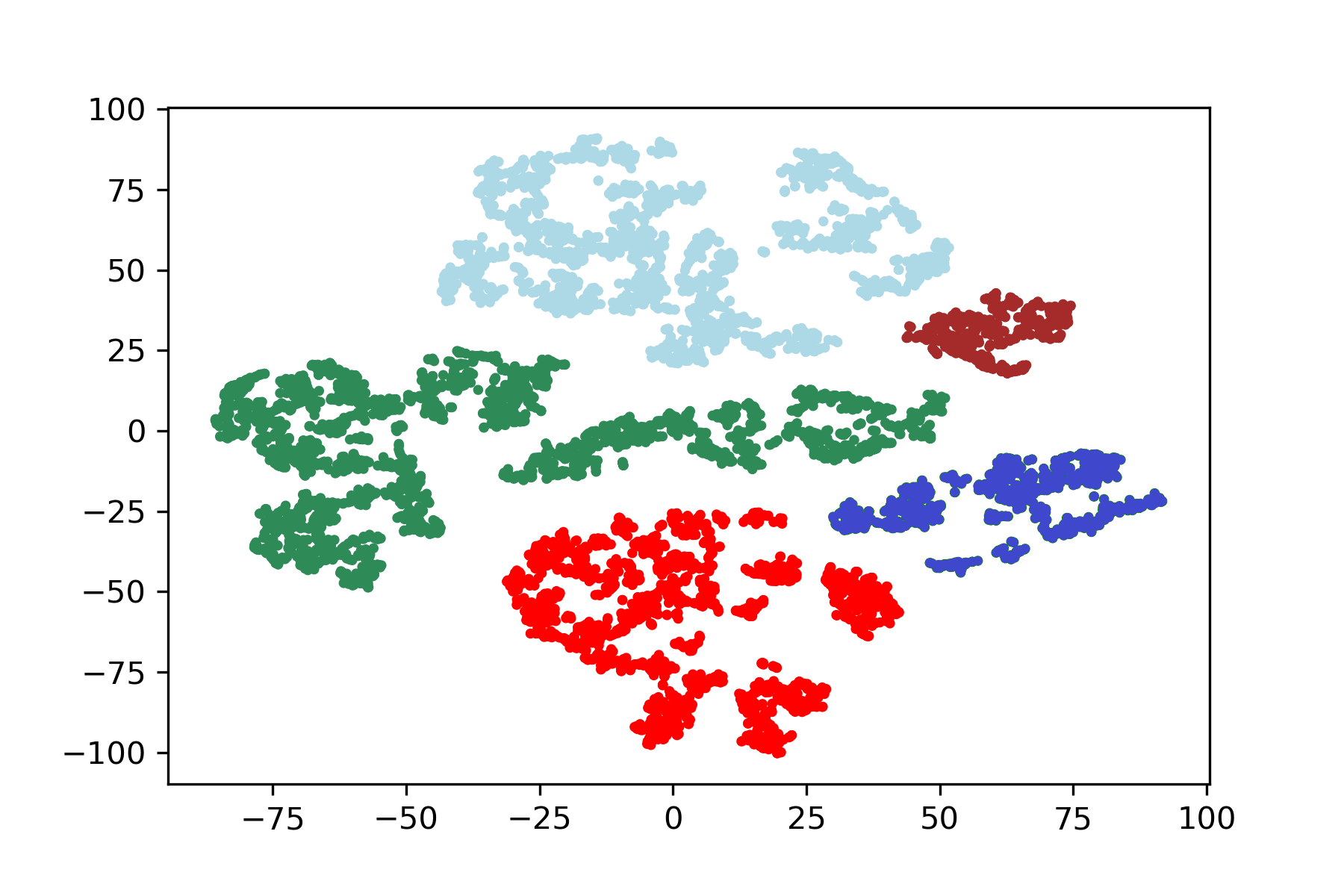}\label{fig:sne2clus}}
     \subfigure[DBSCAN applied to PCA (number of components=4). Red labels indicate noise.]{\includegraphics[width=0.45\textwidth]{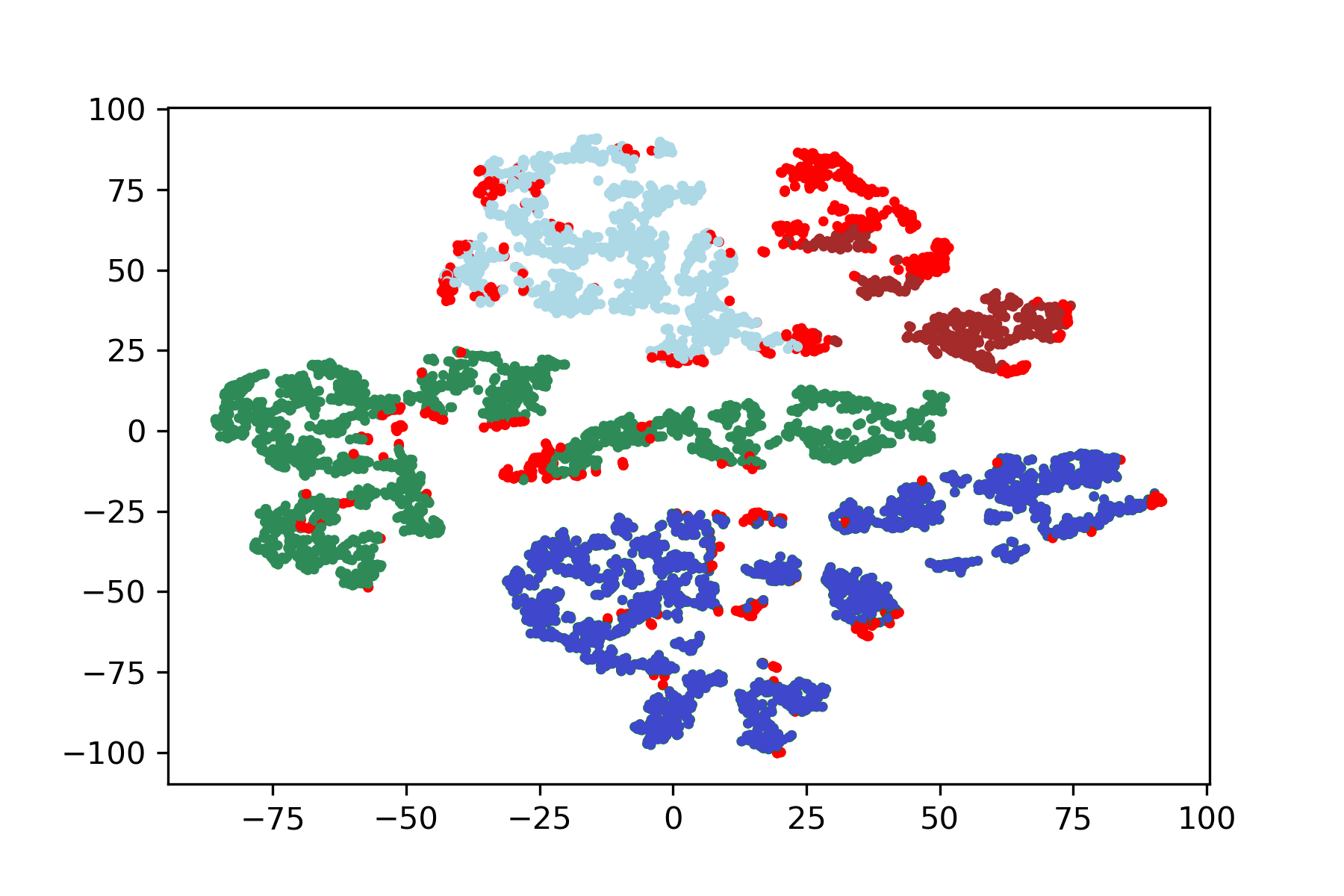}\label{fig:pca4}}
\caption{Results of DBSCAN  on t-SNE embedding (a) and on PCA embedding (b). In (b) the visualization is based on t-SNE, but the labels are obtained by applying DBSCAN on  PCA clustering.}
\label{fig:clust}
\end{figure}

According to Fig. \ref{fig:dimred}, we find the  embedding produced by t-SNE to be the most promising choice to perform clustering. The results of DBSCAN with $\epsilon$ = 9 and $minNeighbors$ = 25 are shown in Fig. \ref{fig:sne2clus}, where five clusters are obtained. Fine-tuning the parameters of DBSCAN, especially, setting $\epsilon$ = 9.6 yields a clustering exactly equivalent to the ground-truth solution. Whereas for a wide range of parameters we obtain the five clusters. These five clusters are consistent with the three ground-truth clusters, i.e., none of them is included in more than one ground-truth clusters. This implies that our solution provides a finer and more detailed representation of the data. It is worth mentioning that the labels we obtain from explicit rule-based approach might not describe the real clusters at a sufficient level, i.e., there might exist finer clusters, especially when dealing with complex scenarios. One may use a hierarchical variant of DBSCAN \cite{DKEChehreghaniAC08} to produce more refined clusters, which can help the domain expert to find and analyze these scenarios in more detail and investigate if we need to expand our scenario catalog with more new scenario classes or keep merging those sub-clusters into a larger scenario class.

We note that t-SNE was originally developed for visualisation and it may sometimes produce misleading results \cite{wattenberg2016how}. However,  there are cases that t-SNE produces a satisfactory embedding for clustering \cite{linderman2019clustering} as in our case.

\begin{figure}[t]
  \begin{center}
    \includegraphics[width=0.45\textwidth]{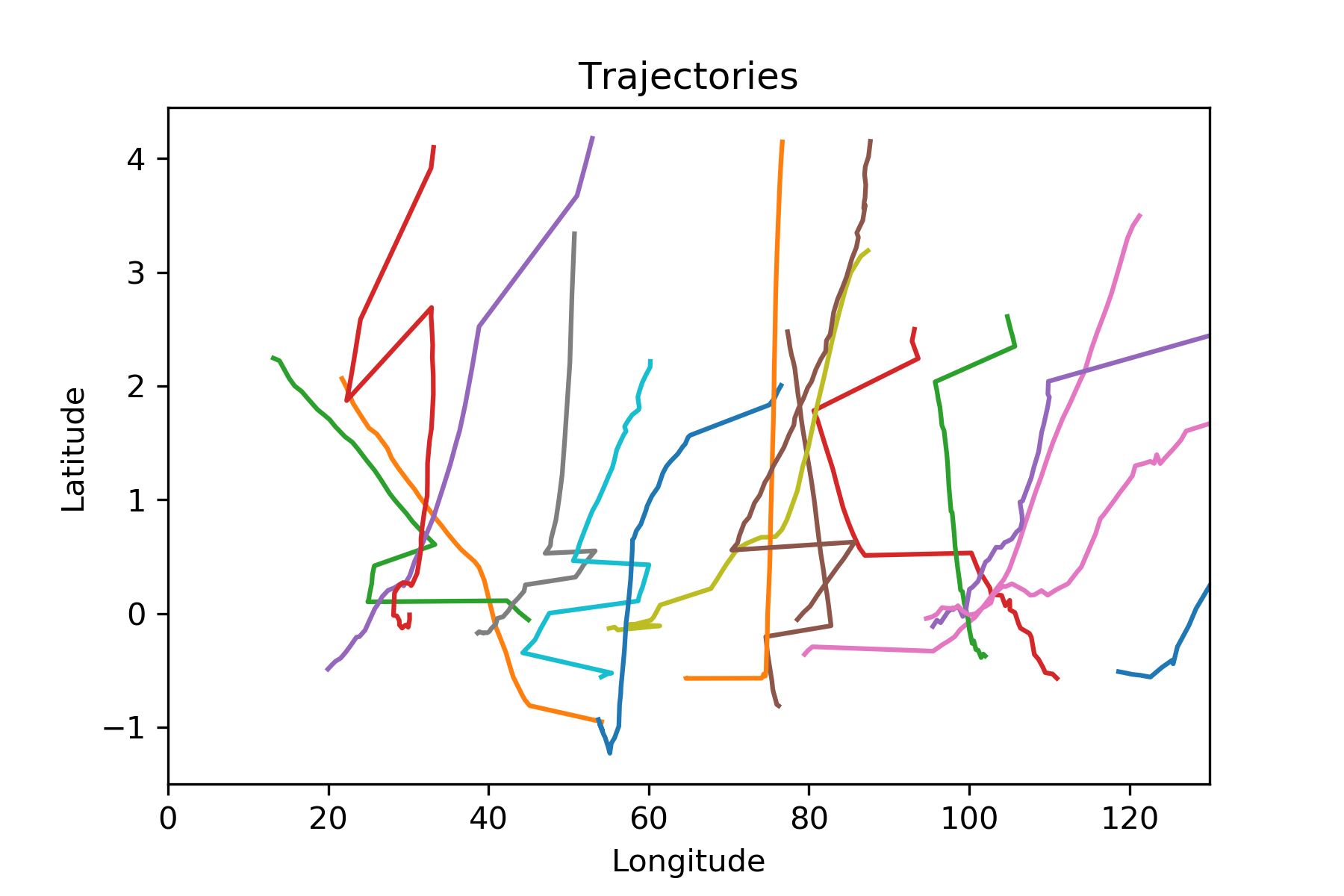}
  \caption{Trajectories with a high probability of being an outlier, obtained by analysis of the autoencoder reconstruction loss.}
  \label{fig:OutlierDetection}
  \end{center}
\end{figure}

In Fig. \ref{fig:pca4}, we apply PCA with four principal components (that cover 75\% of variance) and then apply the clustering method. As it is impossible to plot results in four dimensions, a two-dimensional representation of the trajectories obtained from t-SNE is used to plot the results. While K-means does not produce meaningful results in this four-dimensional space (we assume there is no spherical distribution expected by K-means), more reasonable results are achieved when applying DBSCAN instead. This can be seen in Fig. \ref{fig:pca4}, where four clusters are found. DBSCAN can extract complex and elongated clusters via establishing transitive relations, similar to minimax distance measures \cite{Chehreghani17ICDM,Chehreghani20MLJ} used in \cite{HoseiniRC21}.
It is important to note that some points shown in red are labeled as noise. 

\subsection{Processing and Detection of Outliers}
Finally, we investigate the use of large reconstruction loss in autoencoder in order to detect anomalies of trajectories.
Fig. \ref{fig:OutlierDetection} illustrates the trajectories with high reconstruction loss within the studied set of 2000 cut-ins, say by having a threshold for max reconstruction (relative) error, as discussed in the previous section. We observe that most of the anomaly detected cut-ins are from decelerated cut-ins, which are a minority group w.r.t. all of the $2000$ investigated cut-ins. Also, most of the anomalies (around one percent of the whole set with a high probability to be an outlier) are due to high jumps in the relative longitudinal distance between the ego and the detected surrounding car. This, in general, could be due to various reasons: anomalies in sensing reads of the surrounding object, sudden changes of the two drivers at the same time that can cause considerable changes in the measured relative distance, etc. Note that some jumps in longitudinal/lateral reads could be due to  switching of the detected side of the surrounding cars, detected by the ego car sensor systems. Sensors calculate the relative distance based on the distance of the ego to the mid-point of the closest side of the adjacent cars. However, this side could switch when a car passes the ego car which leads to some jumps in the sensor readings.

A combination of different reasons could also be the root cause.
Outlier detection can provide valuable information, even if we cannot precisely pinpoint the cause. It can be used to improve the quality of the original trajectory dataset, after some more detailed investigations of these anomalies are performed.

\section{Conclusion}
We developed a generic framework for generation and analysis of  driving scenario trajectories based on modern deep neural network models. For trajectory generation, we studied the two generative models AE-GAN (with AE-WGAN-GP extension) and RC-GAN. We adapted them  adequately to handle trajectories with variable length via proper batching the trajectories and integrating a neural component to learn the trajectory lengths.
We also studied in detail the evaluation of the generated trajectories and elaborated several metrics accordingly.

In the following, we studied exploratory analysis of the latent representation from the recurrent autoencoder in a consistent way. In particular, we studied clustering and outlier detection mechanisms based the  output of the trained recurrent autoencoder, where both of them demonstrate promising results.

The proposed framework can be extended in various ways as future work. i)
One direction could be a more sophisticated adjustment of the hyperparameters
of the proposed models with more elegant techniques, rather than the simple grid search used in this work.
ii) AE-GAN is not a conditional model. Hence, to train the length estimator, we  collected ground-truth labels for the length of each encoded trajectory. Then, these labels can be used as a condition to train a similar conditional model. iii)
Considering more features apart from only lateral and longitudinal positions could be possibly helpful for more complex scenarios.

\section*{Acknowledgement}
The work of Morteza Haghir Chehreghani was partially supported by the Wallenberg AI, Autonomous Systems and Software Program (WASP) funded by the Knut and Alice Wallenberg Foundation.
We would like to acknowledge Volvo Cars for providing the data and the computational resources.
We thank Viktor Wänerlöv and Rune Suhr from Volvo Cars, Scenario Analysis team, who helped us throughout the project.

\bibliographystyle{plain}
\bibliography{ref}

\begin{thebibliography}{10}

\bibitem{arnelid2019recurrent}
Henrik Arnelid, Edvin~Listo Zec, and Nasser Mohammadiha.
\newblock Recurrent conditional generative adversarial networks for autonomous
  driving sensor modelling.
\newblock In {\em 2019 IEEE Intelligent Transportation Systems Conference
  (ITSC)}, pages 1613--1618. IEEE, 2019.

\bibitem{Chehreghani16}
Morteza~Haghir Chehreghani.
\newblock Adaptive trajectory analysis of replicator dynamics for data
  clustering.
\newblock {\em Mach. Learn.}, 104(2-3):271--289, 2016.

\bibitem{DKEChehreghaniAC08}
Morteza~Haghir Chehreghani, Hassan Abolhassani, and Mostafa~Haghir Chehreghani.
\newblock Improving density-based methods for hierarchical clustering of web
  pages.
\newblock {\em Data Knowl. Eng.}, 67(1):30--50, 2008.

\bibitem{ChehreghaniRLC07}
Mostafa~Haghir Chehreghani, Masoud Rahgozar, Caro Lucas, and Morteza~Haghir
  Chehreghani.
\newblock A heuristic algorithm for clustering rooted ordered trees.
\newblock {\em Intell. Data Anal.}, 11(4):355--376, 2007.

\bibitem{cho2014learning}
Kyunghyun Cho, Bart Van~Merri{\"e}nboer, Caglar Gulcehre, Dzmitry Bahdanau,
  Fethi Bougares, Holger Schwenk, and Yoshua Bengio.
\newblock Learning phrase representations using rnn encoder-decoder for
  statistical machine translation.
\newblock {\em arXiv preprint arXiv:1406.1078}, 2014.

\bibitem{NIPS2019_8682}
Cyprien de~Masson~d'Autume, Shakir Mohamed, Mihaela Rosca, and Jack Rae.
\newblock Training language gans from scratch.
\newblock In {\em Advances in Neural Information Processing Systems 32}, pages
  4300--4311. 2019.

\bibitem{DemetriouARC20}
Andreas Demetriou, Henrik Allsv{\aa}g, Sadegh Rahrovani, and Morteza~Haghir
  Chehreghani.
\newblock Generation of driving scenario trajectories with generative
  adversarial networks.
\newblock In {\em 23rd {IEEE} International Conference on Intelligent
  Transportation Systems, {ITSC} 2020}, pages 1--6, 2020.

\bibitem{ding2018new}
Wenhao Ding, Wenshuo Wang, and Ding Zhao.
\newblock A new multi-vehicle trajectory generator to simulate
  vehicle-to-vehicle encounters.
\newblock {\em arXiv preprint arXiv:1809.05680}, 2018.

\bibitem{donahue2018adversarial}
David Donahue and Anna Rumshisky.
\newblock Adversarial text generation without reinforcement learning.
\newblock {\em arXiv preprint arXiv:1810.06640}, 2018.

\bibitem{esteban2017real}
Crist{\'o}bal Esteban, Stephanie~L Hyland, and Gunnar R{\"a}tsch.
\newblock Real-valued (medical) time series generation with recurrent
  conditional gans.
\newblock {\em arXiv preprint arXiv:1706.02633}, 2017.

\bibitem{ester1996density}
Martin Ester, Hans-Peter Kriegel, J{\"o}rg Sander, Xiaowei Xu, et~al.
\newblock A density-based algorithm for discovering clusters in large spatial
  databases with noise.
\newblock In {\em Kdd}, volume~96, pages 226--231, 1996.

\bibitem{DTW2018}
Omer Gold and Micha Sharir.
\newblock Dynamic time warping and geometric edit distance: Breaking the
  quadratic barrier.
\newblock {\em ACM Trans. Algorithms}, 14(4), 2018.

\bibitem{goodfellow2014generative}
Ian Goodfellow, Jean Pouget-Abadie, Mehdi Mirza, Bing Xu, David Warde-Farley,
  Sherjil Ozair, Aaron Courville, and Yoshua Bengio.
\newblock Generative adversarial nets.
\newblock In {\em Advances in neural information processing systems}, pages
  2672--2680, 2014.

\bibitem{GulrajaniAADC17}
Ishaan Gulrajani, Faruk Ahmed, Martin Arjovsky, Vincent Dumoulin, and Aaron~C.
  Courville.
\newblock Improved training of wasserstein gans.
\newblock In {\em Advances in Neural Information Processing Systems 30 (NIPS)},
  pages 5767--5777, 2017.

\bibitem{Chehreghani17AAAI}
Morteza Haghir~Chehreghani.
\newblock Classification with minimax distance measures.
\newblock In {\em Thirty-First AAAI Conference on Artificial Intelligence
  (AAAI)}, pages 1784--1790, 2017.

\bibitem{Chehreghani17ICDM}
Morteza Haghir~Chehreghani.
\newblock Efficient computation of pairwise minimax distance measures.
\newblock In {\em 2017 IEEE International Conference on Data Mining (ICDM)},
  pages 799--804, 2017.

\bibitem{Chehreghani20MLJ}
Morteza Haghir~Chehreghani.
\newblock Unsupervised representation learning with minimax distance measures.
\newblock {\em Machine Learning}, 109(11):2063--2097, 2020.

\bibitem{he2016deep}
Kaiming He, Xiangyu Zhang, Shaoqing Ren, and Jian Sun.
\newblock Deep residual learning for image recognition.
\newblock In {\em Proceedings of the IEEE conference on computer vision and
  pattern recognition}, pages 770--778, 2016.

\bibitem{HoseiniRC21}
Fazeleh~Sadat Hoseini, Sadegh Rahrovani, and Morteza~Haghir Chehreghani.
\newblock Vehicle motion trajectories clustering via embedding transitive
  relations.
\newblock In {\em 24th {IEEE} International Intelligent Transportation Systems
  Conference, {ITSC}}, pages 1314--1321. {IEEE}, 2021.

\bibitem{HwangJY19}
Uiwon Hwang, Dahuin Jung, and Sungroh Yoon.
\newblock Hexagan: Generative adversarial nets for real world classification.
\newblock In Kamalika Chaudhuri and Ruslan Salakhutdinov, editors, {\em 36th
  International Conference on Machine Learning, {ICML}}, volume~97, pages
  2921--2930, 2019.

\bibitem{kalra2016driving}
Nidhi Kalra and Susan~M Paddock.
\newblock Driving to safety: How many miles of driving would it take to
  demonstrate autonomous vehicle reliability?
\newblock {\em Transportation Research Part A: Policy and Practice},
  94:182--193, 2016.

\bibitem{kim2016testing}
Baekgyu Kim, Yusuke Kashiba, Siyuan Dai, and Shinichi Shiraishi.
\newblock Testing autonomous vehicle software in the virtual prototyping
  environment.
\newblock {\em IEEE Embedded Systems Letters}, 9(1):5--8, 2016.

\bibitem{NIPS2019_8315}
Vladimir~V. Kniaz, Vladimir Knyaz, and Fabio Remondino.
\newblock The point where reality meets fantasy: Mixed adversarial generators
  for image splice detection.
\newblock In {\em Advances in Neural Information Processing Systems 32}. 2019.

\bibitem{krajewski2018data}
Robert Krajewski, Tobias Moers, Dominik Nerger, and Lutz Eckstein.
\newblock Data-driven maneuver modeling using generative adversarial networks
  and variational autoencoders for safety validation of highly automated
  vehicles.
\newblock In {\em 2018 21st International Conference on Intelligent
  Transportation Systems (ITSC)}, pages 2383--2390. IEEE, 2018.

\bibitem{KurachLZMG19}
Karol Kurach, Mario Lucic, Xiaohua Zhai, Marcin Michalski, and Sylvain Gelly.
\newblock A large-scale study on regularization and normalization in gans.
\newblock In Kamalika Chaudhuri and Ruslan Salakhutdinov, editors, {\em 36th
  International Conference on Machine Learning, {ICML}}, volume~97, pages
  3581--3590, 2019.

\bibitem{li2006coarse}
Xi~Li, Weiming Hu, and Wei Hu.
\newblock A coarse-to-fine strategy for vehicle motion trajectory clustering.
\newblock In {\em 18th International conference on pattern recognition
  (ICPR'06)}, volume~1, pages 591--594. IEEE, 2006.

\bibitem{liao2005clustering}
T~Warren Liao.
\newblock Clustering of time series data—a survey.
\newblock {\em Pattern recognition}, 38(11):1857--1874, 2005.

\bibitem{linderman2019clustering}
George~C Linderman and Stefan Steinerberger.
\newblock Clustering with t-sne, provably.
\newblock {\em SIAM Journal on Mathematics of Data Science}, 1(2):313--332,
  2019.

\bibitem{liu2019driving}
Shiwen Liu, Kan Zheng, Long Zhao, and Pingzhi Fan.
\newblock A driving intention prediction method based on hidden markov model
  for autonomous driving.
\newblock {\em arXiv preprint arXiv:1902.09068}, 2019.

\bibitem{lucic2018gans}
Mario Lucic, Karol Kurach, Marcin Michalski, Sylvain Gelly, and Olivier
  Bousquet.
\newblock Are gans created equal? a large-scale study.
\newblock In {\em Advances in neural information processing systems}, pages
  700--709, 2018.

\bibitem{maaten2008visualizing}
Laurens van~der Maaten and Geoffrey Hinton.
\newblock Visualizing data using t-sne.
\newblock {\em Journal of machine learning research}, 9(Nov):2579--2605, 2008.

\bibitem{malhotra2017timenet}
Pankaj Malhotra, Vishnu TV, Lovekesh Vig, Puneet Agarwal, and Gautam Shroff.
\newblock Timenet: Pre-trained deep recurrent neural network for time series
  classification.
\newblock {\em arXiv preprint arXiv:1706.08838}, 2017.

\bibitem{martinsson2018clustering}
John Martinsson, Nasser Mohammadiha, and Alexander Schliep.
\newblock Clustering vehicle maneuver trajectories using mixtures of hidden
  markov models.
\newblock In {\em 2018 21st International Conference on Intelligent
  Transportation Systems (ITSC)}, pages 3698--3705. IEEE, 2018.

\bibitem{mogren2016c}
Olof Mogren.
\newblock C-rnn-gan: Continuous recurrent neural networks with adversarial
  training.
\newblock {\em arXiv preprint arXiv:1611.09904}, 2016.

\bibitem{nguyen2017m}
Minh Nguyen, Sanjay Purushotham, Hien To, and Cyrus Shahabi.
\newblock m-tsne: A framework for visualizing high-dimensional multivariate
  time series.
\newblock {\em arXiv preprint arXiv:1708.07942}, 2017.

\bibitem{takano2008recognition}
Wataru Takano, Akihiro Matsushita, Keijiro Iwao, and Yoshihiko Nakamura.
\newblock Recognition of human driving behaviors based on stochastic
  symbolization of time series signal.
\newblock In {\em 2008 IEEE/RSJ International Conference on Intelligent Robots
  and Systems}, pages 167--172. IEEE, 2008.

\bibitem{wang2020clustering}
Wenshuo Wang, Aditya Ramesh, Jiacheng Zhu, Jie Li, and Ding Zhao.
\newblock Clustering driving encounter scenarios using connected vehicle
  trajectories.
\newblock {\em IEEE Transactions on Intelligent Vehicles}, 2020.

\bibitem{Wang_2018}
Wenshuo Wang and Ding Zhao.
\newblock Extracting traffic primitives directly from naturalistically logged
  data for self-driving applications.
\newblock {\em IEEE Robotics and Automation Letters}, 3(2):1223–1229, 2018.

\bibitem{wattenberg2016how}
Martin Wattenberg, Fernanda Viégas, and Ian Johnson.
\newblock How to use t-sne effectively.
\newblock {\em Distill}, 2016.

\bibitem{ZhangXLZWHM19}
Han Zhang, Tao Xu, Hongsheng Li, Shaoting Zhang, Xiaogang Wang, Xiaolei Huang,
  and Dimitris~N. Metaxas.
\newblock Stackgan++: Realistic image synthesis with stacked generative
  adversarial networks.
\newblock {\em {IEEE} Trans. Pattern Anal. Mach. Intell.}, 41(8):1947--1962,
  2019.

\bibitem{zhao2017trafficnet}
Ding Zhao, Yaohui Guo, and Yunhan~Jack Jia.
\newblock Trafficnet: An open naturalistic driving scenario library.
\newblock In {\em 2017 IEEE 20th International Conference on Intelligent
  Transportation Systems (ITSC)}, pages 1--8. IEEE, 2017.

\end{thebibliography}

\end{document}